\documentclass{article}

\usepackage{tcolorbox}
\usepackage{listings}
\usepackage{verbatim}
\usepackage[preprint]{neurips_2025}
\usepackage[utf8]{inputenc} 
\usepackage[T1]{fontenc}    
\usepackage{hyperref}       
\usepackage{url}            
\usepackage{booktabs}       
\usepackage{amsfonts}       
\usepackage{nicefrac}       
\usepackage{microtype}      
\usepackage{xcolor}         
\usepackage{pifont} 
\usepackage{graphicx}
\usepackage{amsthm}
\usepackage{amsmath}
\usepackage{subcaption}
\usepackage{paralist}
\usepackage{cases}
\usepackage{booktabs}
\usepackage{threeparttable}
\usepackage{epstopdf}
\usepackage{upgreek}
\usepackage{endnotes}
\usepackage{etoolbox}
\usepackage{algpseudocode}
\usepackage{xspace}
\usepackage{array}
\usepackage{enumitem}
\usepackage{balance}
\usepackage{multirow}
\usepackage{xcolor}
\usepackage{wrapfig}
\usepackage{enumitem}
\usepackage{tabularx}
\usepackage{float}
\usepackage{svg}
\usepackage{subcaption}
\usepackage[ruled,linesnumbered]{algorithm2e}
\usepackage{stfloats}
\usepackage{afterpage}
\usepackage{amsthm}
\usepackage[absolute,overlay]{textpos}

\theoremstyle{definition}
\newtheorem{definition}{Definition}
\newtheorem{problem}{Problem}

\definecolor{deepgreen}{RGB}{34,139,34} 

\newcommand{\red}[1]{{\color{red}{#1}}}
\newcommand{\green}[1]{{\color{deepgreen}{#1}}}

\newcommand{\ie}{\emph{i.e.,}\xspace}
\newcommand{\eg}{\emph{e.g.,}\xspace}

\lstdefinelanguage{json}{
    basicstyle=\ttfamily\small,
    showstringspaces=false,
    breaklines=true,
    frame=lines,
    backgroundcolor=\color{gray!10},
    string=[s]{"}{"},
    comment=[l]{:\ "},
    morecomment=[l]{//},
    keywordstyle=\color{blue},
    stringstyle=\color{teal},
    commentstyle=\color{gray},
    moredelim=[l][\color{black}]{\ },
    morestring=[b]"
}

\title{USTBench: Benchmarking and Dissecting Spatiotemporal Reasoning of LLMs as Urban Agents}

\author{%
  Siqi Lai, Yansong Ning, Zirui Yuan, Zhixi Chen, Hao Liu \\
  The Hong Kong University of Science and Technology (Guangzhou)\\
  \texttt{slai125@connect.hkust-gz.edu.cn} \\
}

\begin{document}

\maketitle

\begin{abstract}
    Large language models (LLMs) have shown emerging potential in spatiotemporal reasoning, making them promising candidates for building urban agents that support diverse urban downstream applications.
    Despite these benefits, existing studies primarily focus on evaluating urban LLM agent on outcome-level metrics (e.g., prediction accuracy, traffic efficiency), offering limited insight into their underlying reasoning processes.
    As a result, the strengths and limitations of urban LLM agents in spatiotemporal reasoning remain poorly understood.
    To this end, we introduce \textbf{USTBench}, the first benchmark to evaluate LLMs’ spatiotemporal reasoning abilities as urban agents across four decomposed dimensions: \textit{spatiotemporal understanding}, \textit{forecasting}, \textit{planning}, and \textit{reflection with feedback}.
    Specifically, USTBench supports five diverse urban decision-making and four spatiotemporal prediction tasks, all running within our constructed interactive city environment \textit{UAgentEnv}.
    The benchmark includes 62,466 structured QA pairs for process-level evaluation and standardized end-to-end task assessments, enabling fine-grained diagnostics and broad task-level comparison across diverse urban scenarios.
    Through extensive evaluation of thirteen leading LLMs, we reveal that although LLMs show promising potential across various urban downstream tasks, they still struggle in long-horizon planning and reflective adaptation in dynamic urban contexts. 
    Notably, recent advanced reasoning models (\eg DeepSeek-R1) trained on general logic or mathematical problems do not consistently outperform non-reasoning LLMs. This discrepancy highlights the need for domain-specialized adaptation methods to enhance urban spatiotemporal reasoning.
    Overall, USTBench provides a foundation to build more adaptive and effective LLM-based urban agents and broad smart city applications. Our project is available at \url{https://github.com/usail-hkust/USTBench}.
\end{abstract}

\vspace{-15pt}
\section{Introduction}\label{sec:intro}

Urban systems are inherently complex and dynamic, characterized by continuous fluctuations across space and time. By learning from large-scale spatiotemporal data, traditional data-driven methods have achieved progress in prediction and decision support \citep{bibri2017smart, zhao2020go, ullah2020applications, wei2019colight, tian2024air}. However, they often fall short in generalizing to unseen scenarios and providing transparent reasoning for reliable decision-making \citep{li2024urbangpt, lai2023llmlight}. Recently, the advanced large language models (LLMs) (\eg GPT-4o \citep{hurst2024gpt} and DeepSeek-R1 \citep{guo2025deepseek}) have emerged as intelligent urban agents \citep{lai2023llmlight,li2024urbangpt,zhou2024large,ning2025dima,ning2024urbankgent} due to their growing reasoning ability to integrate diverse information, adapt across tasks, and offer detailed interpretation through natural language. 
To fully leverage their potential, it is essential to systematically evaluate LLMs' spatiotemporal reasoning abilities: the capacity to infer spatiotemporal dynamics and interact with evolving urban environments. 
Such evaluation is key to understanding their readiness for real-world urban challenges.

\begin{table}[t]
\centering
\caption{Comparison of LLM benchmarks in urban tasks.}
\renewcommand{\arraystretch}{1.1}
\resizebox{\columnwidth}{!}{
\begin{tabular}{c|l|c|c|c|c|c}
\toprule
\multicolumn{2}{c|}{\textbf{Evaluations}} & \textbf{STBench} \citep{li2024stbench} & \textbf{CityBench} \citep{feng2024citybench} & \textbf{CityGPT} \citep{feng2024citygpt} & \textbf{UrbanPlanBench} \citep{zheng2025urbanplanbench} & \textbf{USTBench} (Ours) \\
\hline
& Spatiotemporal Understanding & \green{\ding{51}} & \green{\ding{51}} & \green{\ding{51}} & \red{\ding{55}} & \green{\ding{51}} \\
Reasoning & Forecasting & \green{\ding{51}} & \green{\ding{51}} & \green{\ding{51}} & \red{\ding{55}} & \green{\ding{51}} \\
Abilities & Planning & \red{\ding{55}} & \green{\ding{51}} & \green{\ding{51}} & \red{\ding{55}} & \green{\ding{51}} \\
& Reflection with Feedback & \red{\ding{55}} & \red{\ding{55}} & \red{\ding{55}} & \red{\ding{55}} & \green{\ding{51}} \\
\hline
Baseline & Non-Reasoning LLM  & \green{\ding{51}} & \green{\ding{51}} & \green{\ding{51}} & \red{\ding{55}} & \green{\ding{51}} \\
LLMs & Reasoning LLM  & \red{\ding{55}} & \red{\ding{55}} & \red{\ding{55}} & \red{\ding{55}} & \green{\ding{51}} \\
\hline
Evaluation & Outcome-Based Metrics & \green{\ding{51}} & \green{\ding{51}} & \green{\ding{51}} & \green{\ding{51}} & \green{\ding{51}} \\
Metrics & Process-Based Metrics & \red{\ding{55}} & \red{\ding{55}} & \red{\ding{55}} & \red{\ding{55}} & \green{\ding{51}} \\
\bottomrule
\end{tabular}}
\label{tab:urban-benchmark-comparison}
\vspace{-20pt}
\end{table}

\begin{wrapfigure}{r}{0.5\textwidth}
    \centering
    \includegraphics[width=0.48\textwidth]{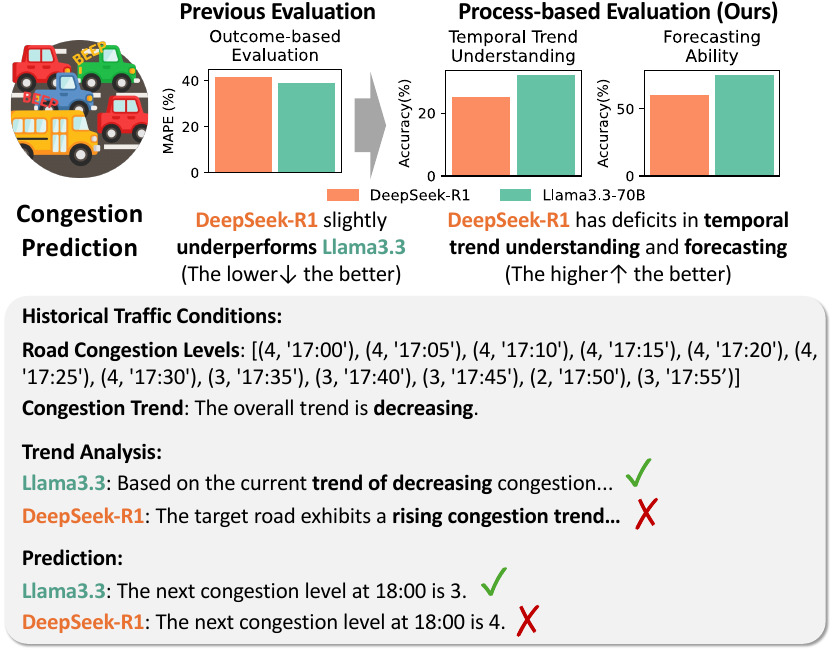}
    \caption{The comparison of outcome-based and process-based evaluations.}
    \label{fig:llama_vs_deepseek}
    \vspace{-10pt}
\end{wrapfigure}


In recent literature, many efforts have been made to evaluate the spatiotemporal reasoning ability of LLMs as urban agents. However, as summarized in Table \ref{tab:urban-benchmark-comparison}, they have two limitations: (1) \textit{Reliance on outcome-based metrics}: solving urban tasks requires multi-step reasoning, yet existing studies \citep{feng2024citygpt, feng2024citybench} only assess outcome-based metrics (\eg prediction accuracy, traffic efficiency), overlooking intermediate reasoning steps. 
Their evaluations may mask critical reasoning deficits. For instance, the reasoning LLM, DeepSeek-R1, typically surpasses non-reasoning models. However, as shown in Figure~\ref{fig:llama_vs_deepseek}, it slightly underperforms Llama3.3 in outcome-based metrics of congestion prediction. Further reasoning process-based analysis reveals that this limitation stems from weaknesses in temporal trend understanding and forecasting. Without reasoning ability evaluations, such discrepancies remain unexplained.
(2) \textit{Overlooking reflection reasoning}: unlike static tasks, urban systems provide real-time and context-rich feedback (\eg shifting traffic patterns), making reflection over past actions essential for agents to adapt to evolving dynamics \cite{wang2024news}. Yet, existing evaluations ignore this process, failing to capture how LLMs improve or adjust over time. These gaps hinder a comprehensive understanding of LLMs' spatiotemporal reasoning in urban tasks. 

To this end, we introduce \textit{USTBench}, the first benchmark designed to systematically evaluate the spatiotemporal reasoning abilities of LLMs as urban agents. 
We first build \textit{UAgentEnv}, an interactive city environment spanning five urban decision-making and four prediction tasks. It enables agents to perceive, interact with, and respond to dynamic urban contexts.
To move beyond outcome-based evaluation, we decompose urban spatiotemporal reasoning abilities into four key processes: spatiotemporal understanding, forecasting, planning, and reflection with feedback. 
Each ability is evaluated through structured question-answering (QA) pairs collected from UAgentEnv. To further explore the interplay between these reasoning processes, we conduct targeted ablation studies, providing diagnostic insights into model strengths and weaknesses.
Combined with end-to-end performance evaluations, this dual-level framework supports both detailed reasoning analysis and standardized downstream task evaluation.

\begin{wrapfigure}{r}{0.5\textwidth}
    \vspace{-13pt}
    \centering
    \includegraphics[width=0.48\textwidth]{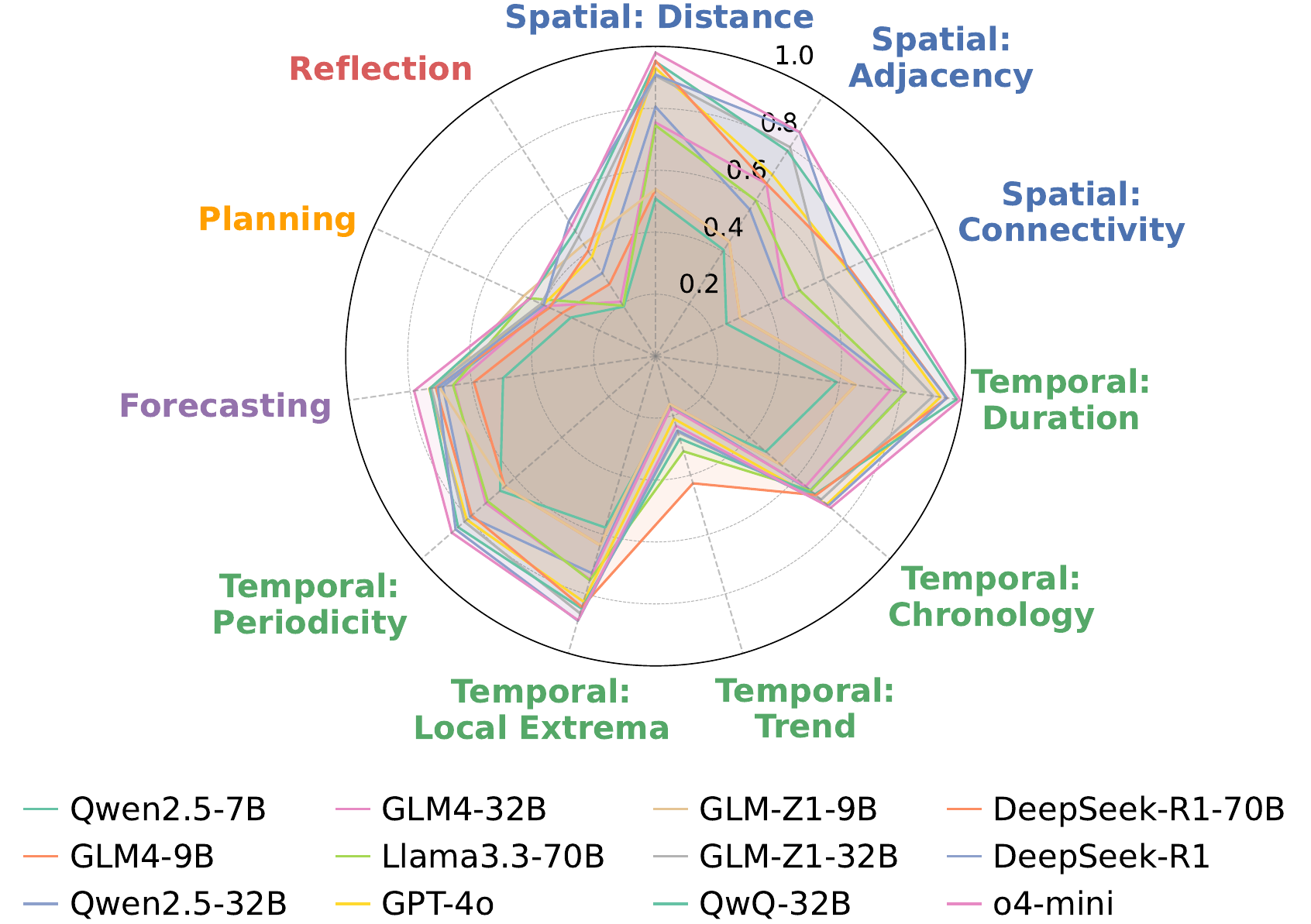}
    \caption{The performance of leading LLMs in urban spatiotemporal reasoning.}
    \label{fig:benchmark_results}
    \vspace{-18pt}
\end{wrapfigure}

Using USTBench, we evaluate thirteen state-of-the-art LLMs, covering both non-reasoning and reasoning models. Our key contributions and findings are summarized as:
(1) We construct USTBench, the first benchmark explicitly designed to evaluate the spatiotemporal reasoning capabilities of LLMs as intelligent urban agents. It combines both fine-grained process-based reasoning evaluation and standardized end-to-end task performance assessment, including 62,466 structured reasoning QA pairs and nine real-world downstream urban tasks.
(2) To support this benchmark, we develop UAgentEnv, an interactive urban environment that enables nuanced benchmark dataset collection and uniform downstream task evaluation across diverse urban scenarios and tasks. 
(3) Despite the promising capability discovered by prior works, our extensive experiments in USTBench reveal several limitations of LLMs. The key findings include: As illustrated in Figure \ref{fig:benchmark_results}, LLMs excels in spatiotemporal understanding and forecasting, but typically struggles in long-term action planning and reflection; Reasoning models trained on general logical or mathematical tasks do not consistently outperform non-reasoning models, underscoring the need for domain-specialized training; Planning emerges as a higher-order ability that LLMs struggle with, which builds upon and extends beyond understanding and forecasting; Reflection is critical for adaptability in dynamic urban contexts, while even the leading reasoning LLMs struggle to learn from environmental feedback.
\vspace{-5pt}
\section{Preliminary}\label{sec:overview}

This work studies the spatiotemporal reasoning abilities of LLMs as urban agents, spanning decision-making and prediction tasks. Here, we provide background knowledge and important definitions.

\begin{problem}
    \textbf{Urban Decision-Making Tasks}: Given an urban environment and a task $t$, an agent operates based on urban observations $o$ with a policy $\pi(o)$ that determines a sequence of decisions $\{a_0, a_1, ..., a_n\}$ to manipulate the environment. The objective is to accomplish a target goal specified by $t$ (\eg optimizing traffic flow by traffic signal control).
\end{problem}

\begin{problem}
    \textbf{Urban Prediction Tasks}: Given historical spatiotemporal observations $\mathbf{o}_i$ of an urban environment, the goal of prediction is to anticipate urban future states $\{s_{i+1}, \ldots, s_{i+\Delta}\}$ over a horizon $\Delta$, where each state $s_i$ captures key indicators (\eg traffic volume) across space and time.
\end{problem}


\begin{definition}
    \textbf{Urban LLM Agents}: An urban LLM agent is a large language model-driven autonomous system designed to operate in dynamic urban environments. Formally, we define the urban environment as $E = \langle S, A, O, T \rangle$, where $S$ states urban state space, $A$ denotes the agent’s action space, $O$ is the observation space, and $T: S \times A \rightarrow S$ is the environment's transition function. At each time step $i$, given a task $t$, the agent receives the current observation $o_i \in O$ (\eg local traffic conditions), along with a history of prior observations $\mathbf{o}_{i-1}$ and actions $\mathbf{a}_{i-1}$. Based on this context, the agent performs reasoning to either (1) execute an action $a_i \in A$ (\eg activate a traffic signal), or (2) generate predictions of future urban states $\{s_{i+1}, \dots, s_{i+\Delta}\} \subset S$ (\eg estimated traffic volume). 
\end{definition}

\begin{definition}
    \textbf{Urban Spatiotemporal Reasoning}: Urban spatiotemporal reasoning is the capability of an LLM-based agent to interpret, act upon, and adapt to urban environments characterized by spatial and temporal dynamics. Formally, given a task $t$ and spatiotemporal observation $o_i$, it involves:
    (1) \textit{Spatiotemporal Understanding} \citep{shi2022stepgame, wang2024tram}: Interpreting urban spatial structures (\eg road network) and temporal patterns (\eg traffic flow shifts) from input observations.
    (2) \textit{Forecasting} \citep{wang2024news}: Reasoning to generate predictions of future urban states $\{s_{i+1}, \ldots, s_{i+\Delta}\}$ based on learned spatiotemporal patterns.
    (3) \textit{Planning} \citep{wang2024survey}: Reasoning to derive control actions $a_i$ that optimize performance objectives within the current and anticipated urban context.
    (4) \textit{Reflection with Feedback}\citep{ji2023towards}: Evaluating the outcomes of decisions or failures of predictions via environmental feedback $f_i$ and updating future reasoning accordingly to improve performance over time.
    \vspace{-5pt}
\end{definition}

\vspace{-5pt}
\section{UAgentEnv: Interactive City Environment}

\vspace{-5pt}
\subsection{Urban Task Suite and Data Sources}

\textbf{Urban Tasks}: UAgentEnv supports nine representative urban tasks across prediction and decision-making. The prediction tasks include next POI prediction \citep{zhao2020go}, congestion forecasting \citep{cheng2018deeptransport}, socio-economic indicator prediction \citep{liu2023knowledge}, and traffic origin-destination (OD) prediction \citep{yuan2024unist}. The decision-making tasks cover traffic signal control \citep{wei2019colight}, POI placement \citep{von2022reinforcement}, route planning \citep{li2021towards}, road planning \citep{zheng2023road}, and urban planning \citep{zheng2023spatial}. Detailed descriptions are provided in Appendix~\ref{subsec:urban_tasks}.

\textbf{Urban Data Collection}: To ensure UAgentEnv reflects real-world urban dynamics, we integrate diverse publicly available datasets across five dimensions:
(1) \textit{Geospatial Data}: We incorporate geospatial data from OpenStreetMap (OSM) \citep{osmplanet}, including points of interest (POIs), areas of interest (AOIs), and road networks.
(2) \textit{Traffic}: Historical traffic flow data from multiple metropolitan areas in China \citep{cheng2018deeptransport, traffic-signal-control} are used to simulate realistic traffic conditions.
(3) \textit{Socio-economy}: We include time-series data on GDP and population trends in Guangzhou (2000–2019), sourced from \citep{wang2022global} and WorldPop \citep{worldpop_hub}, to model urban evolution.
(4) \textit{Human mobility}: Taxi trajectory data from New York City \citep{nyc_tlc_trip_data} capture fine-grained human movement between urban regions.
(5) \textit{POI check-ins}: Check-in records from the FourSquare dataset \citep{wang2023would} capturing individuals transition between POIs over time.

\vspace{-5pt}
\subsection{Unified Urban Agent Framework}\label{subsec:uagent}

We introduce a unified framework for urban tasks, where the urban agent interacts with the environment across nine real-world tasks. To ensure consistency, we standardize task instructions, inputs, outputs, and execution flows using simplistic instructions to showcase LLM basic reasoning abilities.

\begin{figure}[t]
\centering
\includegraphics[width=\textwidth]{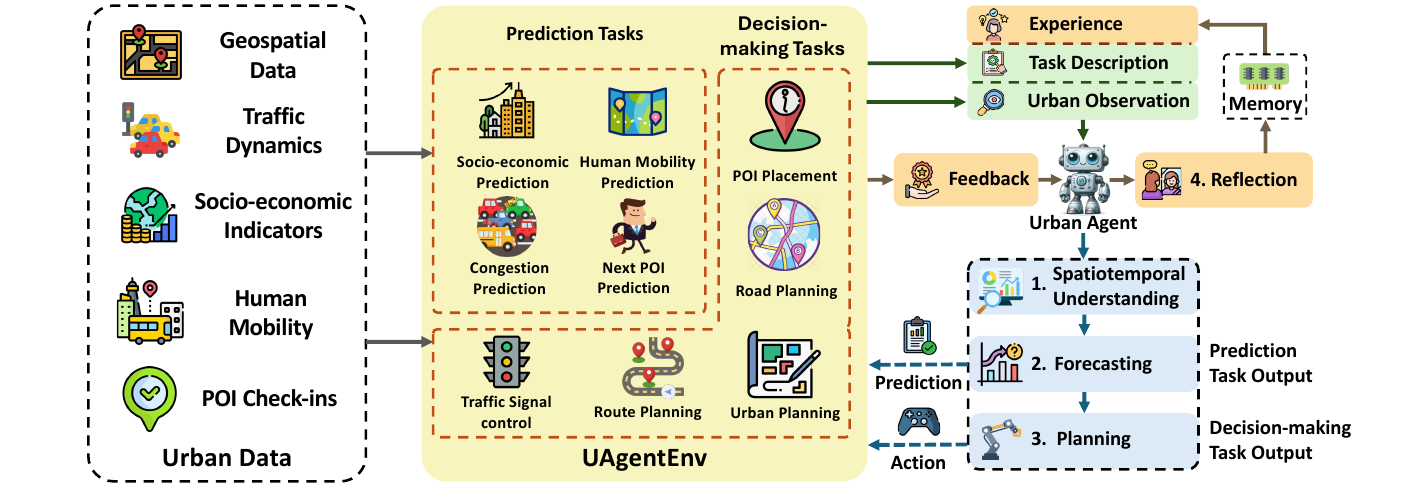}
\centering
\caption{The workflow of UAgentEnv environment.}
\label{fig:simulation}
\vspace{-15pt}
\end{figure}

As shown in Figure~\ref{fig:simulation}, the framework follows a structured pipeline:
(1) Each task provides the agent with a description, data schema, and relevant domain knowledge.
(2) The real-time urban spatiotemporal dynamics (\eg, traffic condition, road configurations) are delivered to the agent as contextual observations.
(3) The LLM agent reasons to solve the task through a modular reasoning workflow comprising spatiotemporal understanding, forecasting, and planning. Then, an action or prediction is generated, aligning with the task objective and relevant past experiences retrieved from memory.
(4) After receiving environmental feedback (\eg improved traffic efficiency) on the previous output, the agent engages in reflection to evaluate its performance and diagnose errors. Then, an informative experience will be generated and stored in the memory to guide future reasoning processes.
The prompt templates are detailed in Appendix~\ref{subsec:agent_prompt}.

\vspace{-5pt}
\section{USTBench: Urban Spatiotemporal Reasoning Benchmark}

\vspace{-5pt}
\subsection{Process-based Spatiotemporal Reasoning Evaluation}
Previous evaluations of LLMs in urban tasks primarily rely on coarse, outcome-based metrics (\eg prediction accuracy or traffic efficiency), which obscure critical reasoning deficits in reasoning abilities. Alternatively, we propose a fine-grained assessment by decomposing spatiotemporal reasoning into four key processes: spatiotemporal understanding, forecasting, planning, and reflection with feedback. These processes are uniformly assessed via structured QA collected from UAgentEnv. USTBench includes 62,466 QA instances across diverse urban scenarios, with performance measured by accuracy. Statistics and examples are shown in Table~\ref{tab:ustbench_tasks}, with details provided in Appendix~\ref{subsubsec:prompt_response}.

\begin{table}[t]
\centering
\renewcommand{\arraystretch}{1.1}
\setlength{\tabcolsep}{1pt}
\caption{The required reasoning abilities of LLM agents and the statistics of our evaluation dataset.}
\resizebox{\columnwidth}{!}{
\begin{tabular}{cl|l|l}
\toprule
\textbf{Reasoning} & \textbf{Category} & \textbf{Observation Example} & \textbf{Question Example} \\
\hline
& Distance & Road length: road1: 126m, road2: 345m... & Rank roads by their distances. \\
& Adjacency & Adjacency: [(region1, region2, 341m), (region1, region3, 125m),...] & Rank the regions by proximity to region 1. \\
Spatiotemporal & Connectivity & Connectivity: [(road1, road2, 121m), (road4, road5, 156m),...] & Is there a path between road 1 and 5? \\
\cline{2-4}
Understanding & Duration & Travel times: route 1: 34min, route 2: 12min,... & Rank the routes by their travel times. \\
(27,000 QAs) & Chronology & Trajectory: [(Shop, 19:34), (Bar, 20:11),...,(Shop, 20:21), (Bar, 20:41),...] & Order the often visited POI pairs. \\
& Trend & Congestion levels: [(2, 8:00), (3, 8:05), (4, 8:10),...] & What is the trend of the congestion? \\
& Local Extrema & Congestion levels: [(4, 10:00),..., (3, 10:00),...,(4, 10:00),...] & Identify high-peak hours in the last 3 days. \\
& Periodicity & Congestion levels: [(4, 9:00),..., (2, 16:00),...,(4, 21:00),...] & Identify the period of the traffic flow pattern. \\
\hline
& Next POI & Trajectory: [(Shop, 19:34, 0m), (Bar, 20:11, 134m),...] & Predict the next POI the user would go. \\
Forecasting & Congestion & Congestion levels: [(2, 8:00), (3, 8:05),...,(4, 8:55)] & Predict the congestion at 9:00. \\
(15,336 QAs) & Socio-economic & Region GDP: [(63M, 2009), (68M, 2010), (71M, 2011),...] & Predict the GDP in 2012. \\
& Traffic-OD & Departure vehicles: [(5, 8:00), (10, 8:05),...,(6, 8:55)] & How many vehicles will depart at 9:00. \\
\hline
& Signal Control & Queues: lane1: 3, lane2: 9,...; Connectivity: [(lane1, lane3, 300m),...]& Which signal should be activated?\\
\multirow{2}{*}{Planning} & POI Placement & Demand: loc1: 12, loc2: 23,...; Distance: loc1: 1021m, loc2: 2033m,... & Where should we placed a new station?\\
\multirow{2}{*}{(15,000 QAs)} & Route Planning & Congestion level: road1: 1, road4: 2,...; Connectivity: [(road1, road2, 126m),...] & Which road should enter next? \\
& Road Planning & Connectivity: [(road1, region2, 134m), (road3, region3, 234m),...] & Which road should be built next? \\
& Urban Planning & Adjacency: [(blank1, residential1, 143m), (blank1, residential2, 345m),...] & Where should we plan a new hospital? \\
\hline
\multirow{2}{*}{Reflection} & Reflection on & \multirow{2}{*}{Ground truth: (4, 10:00); Your prediction: (1, 10:00)} & Your previous prediction is wrong, let’s \\
\multirow{2}{*}{with Feedback} & Forecasting & & review the spatiotemporal data and try again. \\
\cline{2-4}
\multirow{2}{*}{(8,130 QAs)} & Reflection on & \multirow{2}{*}{The facility accessibility is decreased by 12\%.} & Verify if the previous decision \\
& Planning & & and its reasoning are accurate. \\
\bottomrule
\end{tabular}}
\label{tab:ustbench_tasks}
\vspace{-15pt}
\end{table}

\vspace{-5pt}
\subsubsection{Observation Construction}\label{subsubsec:obs_construct}
To systematically evaluate LLMs' reasoning capabilities across diverse urban environments, we collect a variety of urban task scenarios as LLM observations using UAgentEnv. Each observation encodes a snapshot of geospatial structures (\eg road networks) and temporal dynamics (\eg traffic flow patterns) of the city. These serve as the contextual input for LLMs to perform reasoning over specific urban questions. We verbalize spatial observations (\eg road networks) as sparse adjacency matrices with node and edge attributes \citep{chen2024graphwiz}. Temporal observations (\eg trajectories or time-series) are verbalized with attribute values (\eg visited POI or vehicle counts) across discrete time intervals \citep{wang2023would, li2024urbangpt}. Examples of such observations are illustrated in Table~\ref{tab:ustbench_tasks}.

To collect QAs from prediction tasks, we construct observations using a sliding window of length $\Delta$. At each step $i$, we extract real-world historical spatiotemporal dynamics and encode as input observations in QA queries. For QAs constructed from decision-making tasks, we collect observations using a heuristic agent that follows a semi-stochastic policy to interact with the environment for varying urban contexts with diverse spatiotemporal dynamics:
\begin{align}
    \pi_g(o) =
    \begin{cases}
    \arg\max\limits_{a \in A} Q(o, a), & \text{with probability } 1 - \epsilon \\
    \text{random}(A), & \text{with probability } \epsilon
    \end{cases}
\end{align}
where $Q(o, a)$ is a simple utility function that scores the benefit of taking action $a$ in state $s$ (\eg prioritizing the green signal for lanes with the longest queues). To ensure diversity of collected scenarios, we introduce an exploration coefficient $\epsilon \in [0, 1]$, which controls the probability of selecting a random action. This induces diverse decision trajectories resulting in various urban scenarios. At each decision time step $i$, the observed spatiotemporal dynamics is captured and embedded into a corresponding QA instance as agent observations.

\vspace{-5pt}
\subsubsection{Spatiotemporal Understanding QA}\label{subsec:st_data_understand}
Spatiotemporal understanding is the first stage of problem-solving and the foundational ability to interact with urban environments. It involves interpreting spatial relationships among urban entities and identifying temporal patterns in events. To evaluate this ability, we design QA tasks using contextual observations generated in Section~\ref{subsubsec:obs_construct}. Each task challenges the LLM to identify specific spatial relations and temporal patterns shown in the context. In total, the benchmark covers eight well-established types of spatial and temporal patterns, drawing on definitions from prior works \citep{rizvi2024sparc, shi2022stepgame, zheng2023road, cao2024tempo, zhou2019going, vemulapalli2012robust, wang2024tram}. These QAs are designed to assess whether LLMs can accurately extract the correct spatiotemporal pattern shown in observation input.

\textbf{Spatial Understanding}: (1) \textit{Distance} \citep{rizvi2024sparc, li2024reframing}: Evaluating spatial proximity between entities by ranking or comparing distances (\eg closest POI). (2) \textit{Adjacency} \citep{shi2022stepgame, li2024urbangpt}: Identifying whether two entities (\eg roads, regions) are directly connected or adjacent in the spatial layout. (3) \textit{Connectivity} \citep{zheng2023road, zheng2023spatial}: Determining whether a viable path exists between entities in a spatial network.

\textbf{Temporal Understanding}: (1) \textit{Duration} \citep{cao2024tempo, xiong2024large}: Comparing the lengths of time-based events (\eg wait times) (2) \textit{Chronology} \citep{zhou2019going, xiong2024large}: Understanding temporal orderings of events (\eg sequences of POI visits). (3) \textit{Trend} \citep{vemulapalli2012robust, liu2025timecma}: Identifying long-term directional patterns (\eg increasing, decreasing, fluctuating) in time-series data. (4) \textit{Local Extrema} \citep{wang2024tram, fatemi2024test}: Identifying peak or off-peak periods (\eg rush hours) from time-series data fluctuations. (5) \textit{Periodicity} \citep{wang2024tram}: Recognizing the length of periods of patterns shown in time-series data, such as weekly or monthly cycles in urban dynamics.

\vspace{-5pt}
\subsubsection{Forecasting QA}\label{subsec:prediction}
Building on spatiotemporal understanding, forecasting \citep{wang2024news, li2024urbangpt} allows agents to predict future urban states, which is an essential capability not only for prediction tasks but also for modeling action-outcome dependencies in decision-making. To evaluate forecasting as a standalone ability, we construct QAs from the prediction tasks in UAgentEnv. Based on the historical observations $\mathbf{o}_i$, the agent is tasked to predict the future urban state at the next timestep $s_{i+1}$ from a set of candidate options. The ground truth answer is derived from the actual observed value in the real-world data.

\vspace{-5pt}
\subsubsection{Planning QA}\label{subsec:planning}
Planning \citep{wang2024survey, chang2024agentboard} reflects the agent’s ability to reason over spatiotemporal observations and choose actions that optimize long-term urban objectives. Unlike solving static problems (\eg mathematics, web search), planning in urban tasks requires agents to consider benefits over an extended horizon within the complex and evolving environments. To assess this ability, we construct QA instances from five urban decision-making tasks. Given the current spatiotemporal observation $o_i$, the agent is tasked to select an action $a_i \in A$ that best advances the task objective (\eg reducing traffic congestion). The ground-truth answer is computed via an feedback-driven exploratory process, in which all possible future action sequences over a planning horizon $H$ are observed. The action $a^*_i$ that yields the highest expected cumulative feedback reward is selected:
\begin{align}
    a^*_i = \arg\max_{a_i \in A} \max_{a_{i+1}, \ldots, a_{i+H} \in A} \mathbb{E} \left[ \sum_{j=0}^{H} \gamma^j r_{i+j}(a_{i+j}) \mid a_i \right],
\end{align}
where $ r_{i+j}(a_{i+j}) $ is the reward at time $ i+j $ when action $ a_{i+j} $ is taken, and $ \gamma \in [0,1] $ is a discount factor. The expectation is estimated via multiple rollouts to account for environmental stochasticity and ensure the reliability of the ground-truth action.


\vspace{-5pt}
\subsubsection{Reflection with Feedback QA}
Unlike static problem-solving (\eg mathematics), urban systems provide dynamic and context-rich feedback (\eg traffic states at next timesteps). Solving urban tasks requires not only executing accurate actions but also reflecting on previous behaviors to adapt and improve over time \citep{ji2023towards, renze2024self}.

To evaluate this reflective reasoning capability, we construct a dataset in which previous action $a_{i-1}$ or prediction $s_{i-1}$ made by the LLM is paired with environmental feedback $f_i$ observed at the current timestep $i$. The agent is then tasked to assess its prior output and determine whether the decision or prediction was appropriate and, if not, identify the correct answer. This setup tests whether the agent is able to diagnose errors during iterative agent-environment interactions. Such reflection is essential for reasoning in dynamic environments, where the optimal strategy may shift due to evolving spatial or temporal patterns. An agent that can leverage environmental feedback to iteratively refine its reasoning demonstrates a higher level of adaptability and long-term planning competence.

\vspace{-7pt}
\subsection{End-to-End Downstream Task Evaluation}
\vspace{-3pt}

Following previous studies \citep{li2024stbench,feng2024citybench,feng2024citygpt}, we evaluate LLM performance across nine urban tasks in UAgentEnv. Equipped with our urban agent framework, each task is uniformly evaluated using domain-specific metrics. For example, we use Mean Absolute Percentage Error (MAPE) to assess GDP forecasting accuracy over a three-year window in socio-economic prediction. Congestion prediction, which classifies congestion into five levels (0 to 4), is evaluated using accuracy and MAPE. Urban planning performance is assessed based on two criteria: accessibility to service facilities and ecological coverage. For road planning, we measure construction costs and the average travel distance to neighboring regions. The tasks mentioned above are assessed in \ref{subsec:task_evaluation}. Appendix~\ref{sec:detailed_eval} details the evaluation metrics and results of the remaining five tasks.
\vspace{-7pt}
\section{Experiment}

\vspace{-5pt}
\subsection{Baseline Models}
\vspace{-3pt}
We evaluate both non-reasoning LLMs and reasoning models with the same parameter sizes. For non-reasoning LLMs, we include: Qwen2.5 (7B and 32B) \citep{yang2024qwen2}, GLM4 (9B and 32B) \cite{glm2024chatglm}, Llama3.3-70B \cite{grattafiori2024llama}, and GPT-4o \citep{hurst2024gpt}. For reasoning models, we evaluate: DeepSeek-R1-Distill (7B and 70B), DeepSeek-R1 \citep{guo2025deepseek}, QwQ-32B \citep{qwenlm2025qwq32b}, GLM-Z1 (9B and 32B) \citep{glm2024chatglm}, and o4-mini \citep{jaech2024openai}. To contextualize LLM performance in end-to-end downstream tasks, we also compare against traditional domain-specific baselines. Detailed configurations are provided in Appendix~\ref{subsec:baselines}.

\begin{table}[t]
\renewcommand{\arraystretch}{1.1}
\setlength{\tabcolsep}{2pt}
\centering
\caption{Performances on spatiotemporal understanding.}
\resizebox{\columnwidth}{!}{
\begin{tabular}{cccc|c|ccccc|c|c}
\toprule
\multirow{2}{*}{\textbf{Model}} & \multicolumn{4}{c|}{\textbf{Spatial Understanding}} & \multicolumn{6}{c|}{\textbf{Temporal Understanding}} & \multirow{2}{*}{\textbf{Overall}} \\
\cline{2-11}
& \textbf{Distance} & \textbf{Adjacency} & \textbf{Connectivity} & \textbf{Overall} & \textbf{Duration} & \textbf{Chronology} & \textbf{Trend} & \textbf{Local Extrema} & \textbf{Periodicity} & \textbf{Overall} \\
\hline
Random & 0.25 & 0.25 & 0.25 & 0.25 & 0.25 & 0.25 & 0.11 & 0.25 & 0.25 & 0.22 & 0.2344 \\
\hline
\multicolumn{12}{c}{\textbf{Generalist LLMs}} \\
\hline
Qwen2.5-7B & 0.5080 & 0.4070 & 0.2513 & 0.3888 & 0.5902 & 0.4710 & 0.1740 & 0.5767 & 0.6637 & 0.4951 & 0.4552 \\
GLM4-9B & 0.5389 & 0.4400 & 0.2993 & 0.4261 & 0.6522 & 0.5354 & 0.1611 & 0.6362 & 0.6438 & 0.5257 & 0.4883  \\
Qwen2.5-32B & 0.8046 & 0.5623 & 0.4537 & 0.6068 & 0.8136 & 0.6610 & 0.1613 & 0.7303 & 0.7923 & 0.6317 & 0.6224 \\
GLM4-32B & 0.7541 & 0.6618 & 0.4555 & 0.6238 & 0.7662 & 0.6414 & 0.1742 & 0.7513 & 0.7248 & 0.6116 & 0.6162  \\
Llama3.3-70B & 0.7448 & 0.5978 & 0.5113 & 0.6180 & 0.8148 & 0.6630 & 0.3203 & 0.7540 & 0.7163 & 0.6537 & 0.6403 \\
GPT-4o & 0.9295 & 0.6963 & 0.6787 & 0.7681 & 0.9288 & 0.7310 & 0.2110 & 0.8260 & 0.8063 & 0.7006 & 0.7259 \\
\hline
\multicolumn{12}{c}{\textbf{Reasoning LLMs}} \\
\hline
DeepSeek-R1- & \multirow{2}{*}{0.4386} & \multirow{2}{*}{0.0450} & \multirow{2}{*}{0.0337} & \multirow{2}{*}{0.1724} & \multirow{2}{*}{0.5254} & \multirow{2}{*}{0.3890} & \multirow{2}{*}{0.2580} & \multirow{2}{*}{0.2820}  & \multirow{2}{*}{0.3773} & \multirow{2}{*}{0.3663} & \multirow{2}{*}{0.2936}\\
Distill-Qwen-7B & & & & & & & & & & \\
\hline
GLM-Z1-9B & 0.8023 & 0.6929 & 0.5627 & 0.6860 & 0.8126 & 0.67172 & \underline{0.3855} & 0.8271 & 0.8234 & 0.7041 & 0.6973  \\
\hline
QwQ-32B & 0.9508 & 0.7875 & \underline{0.7450} & \underline{0.8278} & \underline{0.9818} & 0.6810 & 0.2777 & 0.8490 & 0.8433 & 0.7266 & 0.7645 \\
GLM-Z1-32B & 0.9053 & 0.8022 & 0.5978 & 0.7684 & 0.9053 & 0.7071 & 0.2555 & 0.8655 & 0.8162 & 0.7099 & 0.7319  \\
\hline
DeepSeek-R1- & \multirow{2}{*}{\underline{0.9528}} & \multirow{2}{*}{0.6618} & \multirow{2}{*}{0.6867} & \multirow{2}{*}{0.7671} & \multirow{2}{*}{0.9500} & \multirow{2}{*}{0.6850} & \multirow{2}{*}{\textbf{0.4280}} & \multirow{2}{*}{0.8440}  & \multirow{2}{*}{0.7850} & \multirow{2}{*}{\underline{0.7384}} & \multirow{2}{*}{0.7492} \\
Distill-Llama-70B & & & & & & & & & & \\
\hline
DeepSeek-R1 & 0.9310 & \textbf{0.8598} & 0.6808 & 0.8239 & 0.9492 & \underline{0.7374} & 0.2502 & \textbf{0.8902} & \underline{0.8540} & 0.7362 & \underline{0.7691}  \\
o4-mini & \textbf{0.9798} & \underline{0.8597} & \textbf{0.7665} & \textbf{0.8687} & \textbf{0.9930} & \textbf{0.7475} & 0.2340 & \underline{0.8884} & \textbf{0.8704} & \textbf{0.7467} & \textbf{0.7924}  \\
\bottomrule
\end{tabular}}
\label{tab:st_understand}
\vspace{-15pt}
\end{table}

\vspace{-5pt}
\subsection{Spatiotemporal Reasoning QA Evaluation}

\vspace{-5pt}
\subsubsection{Spatiotemporal Understanding}\label{subsubsec:st_understand}

\textbf{Overall Results}: 
Table~\ref{tab:st_understand} reports model performance on spatiotemporal understanding. Overall, LLMs excel in interpreting urban spatiotemporal relations and patterns, with all models significantly surpassing the random baseline and reasoning models achieving over 80\% accuracy across multiple abilities. This indicates that broad textual pretraining has embedded transferable priors for urban spatiotemporal reasoning. Additionally, their performance is notably stronger on spatial understanding, reflecting the greater complexity and variability in comprehending temporal dynamics than static spatial structures.
Despite these advantages, LLMs struggle with structured data involving spatial connectivity (\eg road networks), event chronology (\eg POI trajectories), and long-term trends or periodicity in time-series data (\eg traffic flow shifts), which often falls below 70\% accuracy. These challenges likely stem from pretraining predominantly on unstructured text, which limits the models' ability to reason over structured inputs. Moreover, DeepSeek-R1-7B shows a marked drop due to repetition issues \citep{li2023repetition}. Further analysis is provided in Appendix~\ref{subsubsec:repetition}.

\textbf{Non-Reasoning LLMs vs. Reasoning LLMs}: Reasoning models like DeepSeek-R1, QwQ, and GLM-Z1 generally outperform their base models of similar sizes with gains of 7–20\%, highlighting the benefits of reasoning-focused post-training. However, their advantage is not consistent. Firstly, GPT-4o often matches or exceeds models like GLM-Z1-32B and DeepSeek-R1-Distill-70B, suggesting that general post-training on logical and mathematical problems does not always benefit urban spatiotemporal reasoning and may introduce unnecessary complexity. Notably, as stated in Section \ref{sec:intro}, DeepSeek-R1 falls short in temporal trend understanding, exposing its limitations in urban dynamic analysis. In contrast, Llama3.3-70B excels in this ability among non-reasoning models. Interestingly, this strength has been successfully transferred to DeepSeek-R1-70B, which is post-trained on Llama3.3 and excels in this task. This motivates us to further explore targeted enhancement methods, such as domain-adaptive training, to improve these abilities.

\vspace{-7pt}
\subsubsection{Forecasting and Planning}\label{subsubsec:prediction_planning}
\vspace{-3pt}

\textbf{Interplay between Forecasting and Planning}: Table~\ref{tab:problem_solve} summarizes model performance on forecasting and planning. The results show that most LLMs exhibit promising forecasting capabilities, with leading models achieving accuracy above 70\%. In contrast, their performance on planning is substantially lower, revealing that current LLMs struggle to make accurate decisions aligned with long-term objectives.
This disparity highlights the increased complexity of planning and reinforces our claim in Section~\ref{subsec:prediction} that planning is a higher-order ability, dependent on and extending beyond forecasting.
However, in tasks involving long-term temporal trend analysis (\eg congestion and traffic-OD prediction), non-reasoning base models outperform reasoning variants, such as Qwen-2.5 vs. QwQ, and Llama3.3 vs. DeepSeek-R1. This further suggests that the general enhancement of reasoning abilities does not always benefit unique challenges in urban scenarios.

\begin{figure}[t]
\centering
\includegraphics[width=\textwidth]{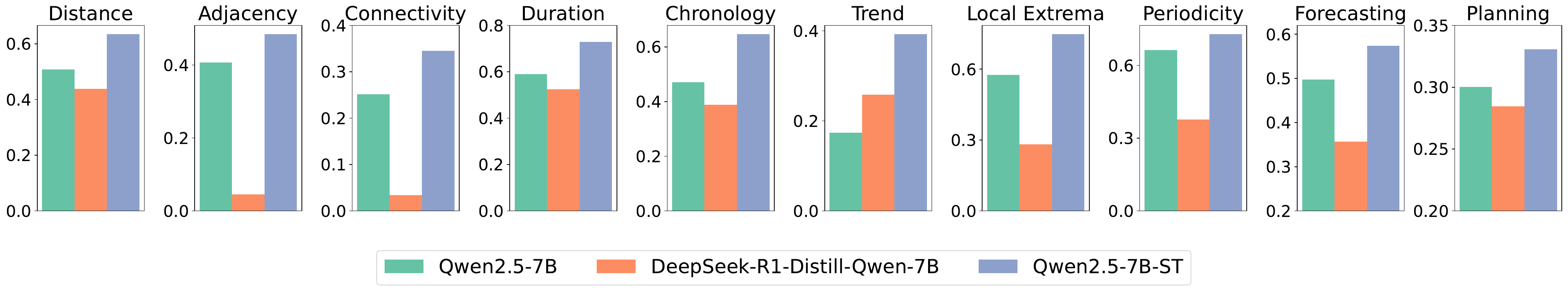}
\centering
\caption{The performance of the model with enhanced spatiotemporal understanding abilities.}
\label{fig:qwen_st}
\vspace{-15pt}
\end{figure}

\textbf{Interplay among Spatiotemporal Understanding, Forecasting, and Planning}\label{sububsec:forecast_plan}:
Models that excel in spatiotemporal understanding generally perform better in forecasting and planning. 
For instance, we observed similar results on Llama3.3 and DeepSeek-R1 as Section \ref{subsubsec:st_understand}, where Llama3.3 not only excels on temporal trend understanding but also in congestion and traffic-OD prediction.
To validate this connection, we post-train Qwen2.5-7B on synthetic spatiotemporal reasoning datasets (the training details are shown in Appendix~\ref{sec:llm_post_training}. As illustrated in Figure \ref{fig:qwen_st}, the fine-tuned model (Qwen2.5-7B-ST) significantly outperforms its base version, confirming the benefit of improved spatiotemporal understanding to downstream reasoning processes. Future research could focus on: (1) improving LLMs' reasoning ability over complex structured spatiotemporal data, and (2) integrating tools and code-based modeling to support more robust spatiotemporal pattern extraction. Both avenues could further advance LLM effectiveness in these two downstream reasoning abilities.

\begin{table}[t]
\renewcommand{\arraystretch}{1.1}
\setlength{\tabcolsep}{2pt}
\centering
\caption{Performance of LLMs in forecasting, planning, and reflection abilities.}
\resizebox{\columnwidth}{!}{
\begin{tabular}{c|cccc|c|ccccc|c|c}
\toprule
\multirow{3}{*}{\textbf{Model}} & \multicolumn{5}{c|}{\textbf{Forecasting}} & \multicolumn{6}{c|}{\textbf{Planning}} & \multirow{3}{*}{\textbf{Reflection}} \\
\cline{2-12}
& \textbf{Next POI} & \textbf{Socio-economic} & \textbf{Congestion} & \textbf{Traffic-OD} & \multirow{2}{*}{\textbf{Overall}} & \textbf{Traffic Signal}	& \textbf{POI} & \textbf{Road} & \textbf{Route} & \textbf{Urban} & \multirow{2}{*}{\textbf{Overall}} & \\
& \textbf{Prediction} & \textbf{Prediction} & \textbf{Prediction} & \textbf{Prediction} & & \textbf{Control} & \textbf{Placement} & \textbf{Planning} & \textbf{Planning} & \textbf{Planning} & & \\
\hline
Random & 0.25 & 0.25 & 0.25 & 0.25 & 0.25 & 0.25 & 0.25 & 0.25 & 0.25 & 0.25 & 0.25 & 0.25 \\
\hline
\multicolumn{13}{c}{\textbf{Generalist LLMs}} \\
\hline
Qwen2.5-7B & 0.4640 & 0.4879 & 0.6463 & 0.3907 & 0.4972 & 0.3760 & 0.2190 & 0.2960 & 0.3490 & 0.2620 & 0.3004 & 0.1899    \\
GLM4-9B & 0.5660 & 0.6456 & 0.6880 & 0.4690 & 0.5922 & 0.3920 & 0.2890 & 0.2930 & 0.4430 & 0.2480 & 0.3330 & 0.2758    \\
Qwen2.5-32B & 0.7473 & 0.8449 & 0.6900 & 0.4930 & 0.6938 & 0.5227 & 0.2047 & 0.3030 & 0.5483 & 0.3430 & 0.3843 & 0.3184    \\
GLM4-32B & 0.7090 & 0.7812 & 0.6730 & 0.4640 & 0.6568 & 0.5550 & 0.2230 & 0.3910 & 0.4740 & 0.3010 & 0.3888 & 0.2087    \\
Llama3.3-70B & 0.7507 & 0.6301 & \textbf{0.7493} & 0.5110 & 0.6603 & 0.5313 & 0.3273 & 0.3563 & \textbf{0.6950} & 0.3413 & \underline{0.4503} & 0.1935    \\
GPT-4o & 0.8280 & 0.9029 & \underline{0.7380} & 0.4840 & \underline{0.7382}  & 0.5620 & 0.2820 & 0.3330 & 0.4780 & 0.3380 & 0.3986 & 0.3802   \\
\hline
\multicolumn{13}{c}{\textbf{Reasoning LLMs}} \\
\hline
DeepSeek-R1- & \multirow{2}{*}{0.4990} & \multirow{2}{*}{0.2488} & \multirow{2}{*}{0.4597} & \multirow{2}{*}{0.2193} & \multirow{2}{*}{0.3567} & \multirow{2}{*}{0.4120} & \multirow{2}{*}{0.1507} & \multirow{2}{*}{0.2000} & \multirow{2}{*}{0.4117} & \multirow{2}{*}{0.2493} & \multirow{2}{*}{0.2847} & \multirow{2}{*}{0.1068}  \\
Distill-Qwen-7B & & & & & & & & & & & & \\
\hline
GLM-Z1-9B & 0.8740 & \underline{0.9159} & 0.6020 & 0.4280 & 0.7050 & 0.5280 & \textbf{0.4990} & 0.3960 & \underline{0.6000} & 0.3160 & \textbf{0.4678} & 0.4293  \\
\hline
QwQ-32B & 0.9153 & 0.9015 & 0.6290 & 0.4987 & 0.7361 & \underline{0.5637} & \underline{0.4897} & 0.3587 & 0.4223 & 0.3997 & 0.4468 & 0.4804  \\
GLM-Z1-32B & \underline{0.9230} & 0.8957 & 0.6690 & 0.4360 & 0.7309 & 0.5520 & 0.4100 & 0.3910 & 0.3660 & 0.3120 & 0.4062 & 0.4597  \\
\hline
DeepSeek-R1- & \multirow{2}{*}{0.8737} & \multirow{2}{*}{0.8691} & \multirow{2}{*}{0.5947} & \multirow{2}{*}{\textbf{0.5280}} & \multirow{2}{*}{0.7164} & \multirow{2}{*}{0.5173} & \multirow{2}{*}{0.3227} & \multirow{2}{*}{0.3703} & \multirow{2}{*}{0.3517} & \multirow{2}{*}{0.3193} & \multirow{2}{*}{0.3763} & \multirow{2}{*}{0.4035}  \\
Distill-Llama-70B & & & & & & & & & & & & \\
\hline
DeepSeek-R1 & 0.8900 & 0.8706 & 0.6020 & 0.4780 & 0.7101 & 0.5160 & 0.3120 & \underline{0.4060} & 0.3120 & \underline{0.4380} & 0.3968 & \textbf{0.5179}  \\
o4-mini & \textbf{0.9320} & \textbf{0.9709} & 0.7360 & \underline{0.5100} & \textbf{0.7872} & \textbf{0.5860} & 0.3420 & \textbf{0.4610} & 0.3840 & \textbf{0.4600 }& 0.4466 & \underline{0.5011}   \\
\bottomrule
\end{tabular}}
\label{tab:problem_solve}
\vspace{-10pt}
\end{table}

\begin{table}[t]
\renewcommand{\arraystretch}{1.1}
\setlength{\tabcolsep}{12pt}
\centering
\caption{Performance on downstream urban tasks. Lower MAPE, Cost, and Distance indicate better performance, while higher Accuracy, Service, and Ecology scores reflect improved outcomes.}
\resizebox{\columnwidth}{!}{
\begin{tabular}{cc|cc|cc|cc}
\toprule
\multirow{2}{*}{\textbf{Model}} & \textbf{Socio-economic Prediction} & \multicolumn{2}{c|}{\textbf{Congestion Prediction}} & \multicolumn{2}{c|}{\textbf{Urban Planning}} & \multicolumn{2}{c}{\textbf{Road Planning}} \\
\cline{2-8}
& MAPE & MAPE & Accuracy & Service & Ecology & Cost & Distance \\
\hline
Classic Method & 7.09\%  & 57.05\%  & 17.18\%  & 0.6100  & 0.4310  & 18.95 & 1.99  \\
\hline
\multicolumn{8}{c}{\textbf{Generalist LLMs}} \\
\hline
Qwen2.5-7B & 34.57\% & 66.19\%  & 40.51\% & 0.5951 & \underline{0.6440} & 20.72 & 1.50 \\
GLM4-9B & 58.43\% & 41.41\% & 54.71\% & 0.6355 & 0.4507 & 20.59 & 1.50	\\
Qwen2.5-32B & 6.00\% & \underline{24.90\%}  & \underline{65.90\%} & 0.6335 & 0.5209 & 20.56 & 1.55 \\
GLM4-32B & 9.41\% & 28.61\% & 63.02\% & 0.6662 & 0.4715 & 18.44 & 1.52 \\
Llama3.3-70B & 10.86\% & 38.88\% & 56.10\% & 0.6561 & 0.5842 & 19.10 & 1.57 \\
\hline
\multicolumn{8}{c}{\textbf{Reasoning LLMs}} \\
\hline
DeepSeek-R1- & \multirow{2}{*}{79.23\%} & \multirow{2}{*}{67.42\%} & \multirow{2}{*}{37.88\%} & \multirow{2}{*}{0.6348} & \multirow{2}{*}{0.6111} & \multirow{2}{*}{20.60} & \multirow{2}{*}{1.47} \\
Distill-Qwen-7B & & & & & & & \\
\hline
GLM-Z1-9B & 11.58\% & 45.87\% & 52.01\% & 0.6443 & 0.5430 & 18.80 & 1.33 \\
\hline
QwQ-32B & 5.64\% & 44.89\% & 52.88\% & \underline{0.6751} & 0.5792 & \textbf{18.40} & 1.77 \\
GLM-Z1-32B & 7.55\% & 47.93\% & 51.22\% & 0.6468 & 0.3965 & 18.57 & 1.87 \\
\hline
DeepSeek-R1- & \multirow{2}{*}{5.94\%} & \multirow{2}{*}{38.78\%} & \multirow{2}{*}{55.50\%} & \multirow{2}{*}{0.6560} & \multirow{2}{*}{0.4711} & \multirow{2}{*}{19.42} & \multirow{2}{*}{\textbf{1.13}} \\
Distill-Llama-70B & & & & & & & \\
\hline
DeepSeek-R1 & \underline{5.24\%} & 41.38\% & 58.75\% & \textbf{0.6858} & \textbf{0.6651} & \underline{18.49} & 1.86 \\
o4-mini & \textbf{4.97\%} & \textbf{15.78\%} & \textbf{75.73\%} & 0.6544 & 0.3863 & 19.60 & \underline{1.23} \\
\bottomrule
\end{tabular}}
\label{tab:task_evaluation}
\vspace{-10pt}
\end{table}

\begin{figure}[t]
\centering
\includegraphics[width=\textwidth]{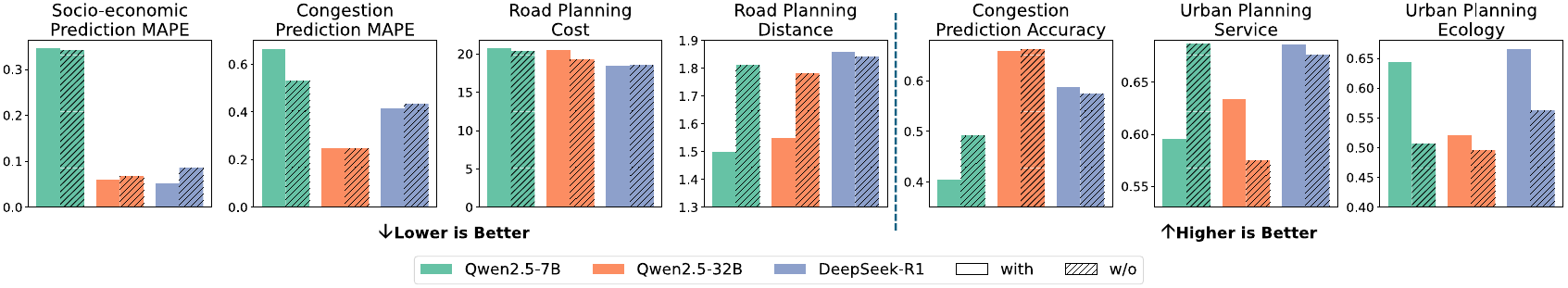}
\centering
\caption{The performance of LLM agents with or without the reflection mechanism.}
\label{fig:wo_reflection}
\vspace{-15pt}
\end{figure}

\vspace{-7pt}
\subsubsection{Reflection with Feedback}

\textbf{Limitations in Reflection Ability}: Table~\ref{tab:problem_solve} shows that most LLMs achieve less than 50\% accuracy on reflection tasks, highlighting a key limitation: current models struggle to incorporate feedback and adapt their reasoning over time. Reflection requires diagnosing errors and generalizing experience to subsequent reasoning and actions, which is a critical ability for long-term adaptation and strategy adjustment in dynamic urban settings. We further analyze how reflection influences downstream task performance and other reasoning processes in Section~\ref{subsec:task_evaluation}.

\vspace{-7pt}
\subsection{Downstream Task Evaluation}\label{subsec:task_evaluation}

\textbf{Overall Results}: Table~\ref{tab:task_evaluation} presents the performance of LLMs on four representative downstream urban tasks. Generally, LLMs outperform traditional domain-specific baselines across prediction and decision-making tasks. Notably, we observe relative performance improvements of up to 337.31\% in forecasting accuracy and 53.48\% in decision outcomes. This reinforces the promise of LLMs as flexible and robust agents for various spatiotemporal tasks in urban scenarios. In particular, models such as Qwen2.5-32B and LLaMA3.3 achieve superior performance compared to their reasoning-augmented variants (\eg QwQ and DeepSeek-R1) on tasks like congestion prediction. This aligns with our earlier findings in Section~\ref{sububsec:forecast_plan}, where process-based evaluations revealed their reasoning bottlenecks in forecasting capabilities.


\textbf{The Impact of Reflection}: To assess the role of reflection, we conducted ablation studies comparing agent performance with and without reflective reasoning. As shown in Figure~\ref{fig:wo_reflection}, its removal led to clear performance drops on DeepSeek-R1, the model with the best reflection ability, highlighting its importance in enhancing adaptability in dynamic urban environments. However, this benefit is less consistent in models with moderate reflection capabilities (\eg Qwen2.5-32B), where reflective outputs do not reliably enhance subsequent forecasting or planning. Conversely, in models with limited reflection ability (\eg Qwen2.5-7B), disabling reflection can even improve performance, suggesting that low-quality reflective content may introduce noise and degrade downstream reasoning.

\vspace{-7pt}
\section{Related Work}
\vspace{-3pt}

\textbf{Urban Agent}: Urban agents have evolved from rule-based systems to RL agents and now LLM-based models. Early systems like SCOOT \citep{hunt1982scoot} and SCATS \citep{lowrie1990scats} used fixed heuristics and sensor data for traffic control. RL approaches, such as CoLight \citep{wei2019colight}, introduced data-driven adaptive learning using graph attention networks. Recently, LLM-based agents, like LLMLight \citep{lai2023llmlight}, UrbanGPT\citep{li2024urbangpt}, and UrbanKGent \citep{ning2024urbankgent}, leverage LLMs for tasks like traffic optimization, spatiotemporal forecasting, and urban knowledge base construction, enabling more flexible and scalable urban intelligence.

\textbf{Spatiotemporal Reasoning}: Spatial reasoning in LLMs has been assessed by understanding spatial relations \citep{mirzaee2021spartqa, mirzaee2022transfer}. Later works introduced multi-hop reasoning tasks \citep{shi2022stepgame} and urban-scale challenges \citep{zhan2025open3dvqa, zhao2025urbanvideo}. Notably, CityEQA \citep{zhao2025cityeqa} extends spatial reasoning to embodied agents navigating city spaces. Temporal reasoning focuses on understanding event sequences and durations. Studies such as \citep{zhou2019going, fatemi2024test, tan2023towards, chu2023timebench} examined commonsense and structured temporal logic. Spatiotemporal reasoning integrates both spatial and temporal dynamics. Recent work \citep{mooney2023towards, yamada2023evaluating, gurnee2023language} analyzed how LLMs interpret spatiotemporal patterns. In urban contexts, benchmarks like STBench \citep{li2024stbench} and CityBench \citep{feng2024citybench} evaluate the reasoning abilities on trajectories and interactions between spatial entities over time.

\textbf{LLM Complex Reasoning}: LLM reasoning has seen rapid advances through chain-of-thought techniques \citep{wei2022chain}, which improves multi-step problem solving by encouraging intermediate reasoning. First, instruction tuning approaches, such as AgentTuning \citep{zeng2023agenttuning}, boost reasoning quality. Recently, post-training strategies adopted by models like OpenAI-o1 \citep{jaech2024openai}, DeepSeek-R1 \citep{guo2025deepseek}, and QwQ \citep{qwenlm2025qwq32b} incorporate RL algorithms like Proximal Policy Optimization (PPO) \citep{schulman2017proximal} and Group Relative Policy Optimization (GRPO) to refine reasoning through reward-guided optimization.

\vspace{-10pt}
\section{Conclusion and Limitations}\label{sec:conclusion}
\vspace{-5pt}

\textbf{Conclusion}: We present USTBench, the first benchmark for systematically evaluating the spatiotemporal reasoning abilities of LLMs as urban agents. Built on the interactive environment UAgentEnv, USTBench supports both fine-grained diagnostics to specific reasoning abilities and end-to-end task evaluations for standardized performance assessment.
Our evaluation of leading LLMs shows that while current models excel in spatiotemporal understanding and forecasting, they struggle with higher-order reasoning tasks, particularly long-term action planning and adaptive reflection. Notably, reasoning models trained on general logic and mathematics do not consistently outperform non-reasoning models in urban-specific tasks, highlighting the need for domain-specific approaches.

\textbf{Limitations}: USTBench primarily focuses on evaluation, while systematic methods for enhancing spatiotemporal reasoning are still underexplored. Additionally, our evaluations are conducted mainly in simulated environments. Although simulations are controllable and scalable, real-world validation and human assessment are essential for urban applications. Future work will explore targeted training approaches and integrate real-world experiments with human evaluations to address these gaps.

\bibliographystyle{plain}
\bibliography{neurips_2025}
\clearpage
\appendix

\section{Limitations}\label{sec:app_lim}
\vspace{-5pt}

\textbf{Exploration on Enhancement Methods}:
USTBench is designed primarily as a comprehensive benchmark to rigorously evaluate the spatiotemporal reasoning abilities of LLMs as urban agents. However, a systematic method for improving these reasoning capabilities remains underexplored. Current LLMs often struggle with complex, long-term planning and adaptive reflection, while USTBench only proposed a simple method with post-training on spatiotemporal understanding. Developing targeted learning approaches, such as domain-specific fine-tuning, reinforcement learning with feedback, or multi-modal integration, will be crucial next steps to enhance LLM performance on spatiotemporal urban reasoning tasks.

\textbf{Real-world Environment Benchmarking}:
While USTBench leverages the UAgentEnv to provide controlled, scalable, and diverse urban scenarios for evaluation, this approach inherently limits exposure to real-world complexity and noise. Simulations may not fully capture unexpected events and rare occurrences reflecting urban dynamics. Moreover, human judgment and expert assessment are critical in real urban planning and management. Future work should incorporate real-world field tests and integrate human feedback to validate and improve LLM performance, ensuring practical applicability and robustness in live urban environments.

\vspace{-5pt}
\section{Confidence Interval of Evaluation}\label{sec:error_bar}
\vspace{-5pt}

In Table \ref{tab:confidence_interval}, we report the confidence intervals for representative models based on three experimental runs with different random seeds.

\begin{table}[h]
\vspace{-10pt}
\setlength{\tabcolsep}{2pt}
\renewcommand{\arraystretch}{1.1}
\centering
\caption{Confidence Interval of Process-based Spatiotemporal Reasoning Evaluation with Accuracy (\%).}
\resizebox{\columnwidth}{!}{
\begin{tabular}{cccc|ccccc|c|c|c}
\toprule
\multirow{2}{*}{\textbf{Model}} & \multicolumn{3}{c|}{\textbf{Spatial Understanding}} & \multicolumn{5}{c|}{\textbf{Temporal Understanding}} & \multirow{2}{*}{\textbf{Forecasting}} & \multirow{2}{*}{\textbf{Planning}} & \multirow{2}{*}{\textbf{Reflection}} \\
\cline{2-9}
& \textbf{Distance} & \textbf{Adjacency} & \textbf{Connectivity} & \textbf{Duration} & \textbf{Chronology} & \textbf{Trend} & \textbf{Local Extrema} & \textbf{Periodicity} & \\
\hline
\multicolumn{11}{c}{\textbf{Non-Reasoning LLMs}} \\
\hline
Qwen2.5-7B & 50.80 ($\pm 1.80)$ & 40.70 ($\pm 0.90$) & 25.13 ($\pm 0.08$) & 59.02 ($\pm 0.21$) & 47.10 ($\pm 0.57$) & 17.40 ($\pm 0.12$) & 57.67 ($\pm 0.67$) & 66.37 ($\pm 0.10$) & 49.72 ($\pm 0.10$) & 30.04 ($\pm 0.58$) & 18.99 ($\pm 0.17$) \\
Qwen2.5-32B & 80.46 ($\pm 0.28$) & 56.23 ($\pm 0.73$) & 45.37 ($\pm 4.10$) & 81.36 ($\pm 0.13$) & 66.10 ($\pm 1.23$) & 16.13 ($\pm 0.03$) & 73.03 ($\pm 0.29$) & 79.23 ($\pm 0.66$) & 69.38 ($\pm 0.23$) & 38.43 ($\pm 5.53$) & 31.84 ($\pm 0.10$) \\
\hline
\multicolumn{11}{c}{\textbf{Reasoning LLMs}} \\
\hline
DeepSeek-R1- & \multirow{2}{*}{43.86 ($\pm 0.38$)} & \multirow{2}{*}{4.50 ($\pm 0.10$)} & \multirow{2}{*}{3.37 ($\pm 0.07$)} & \multirow{2}{*}{52.54 ($\pm 0.38$)} & \multirow{2}{*}{38.90 ($\pm 1.27$)} & \multirow{2}{*}{25.80 ($\pm 0.32$)} & \multirow{2}{*}{28.20 ($\pm 1.41$)} & \multirow{2}{*}{37.73 ($\pm 1.43$)}  & \multirow{2}{*}{35.67 ($\pm 0.46$)} & \multirow{2}{*}{28.47 ($\pm 0.28$)} & \multirow{2}{*}{10.68 ($\pm 0.42$)} \\
Distill-Qwen-7B & & & & & & & & & & & \\
\hline
QwQ-32B & 95.08 ($\pm 3.02$) & 78.75 ($\pm 0.68$) & 74.50 ($\pm 1.01$) & 98.18 ($\pm 1.89$) & 68.10 ($\pm 2.57$) & 27.77 ($\pm 0.11$) & 84.90 ($\pm 1.11$) & 84.33 ($\pm 0.86$) & 73.61 ($\pm 0.41$) & 44.68 ($\pm 1.44$) & 48.04 ($\pm 0.17$) \\
\bottomrule
\end{tabular}}
\label{tab:confidence_interval}
\vspace{-10pt}
\end{table}

\vspace{-5pt}
\section{Runtime Estimation}\label{sec:runtime}
\vspace{-5pt}

The evaluation runtime of an LLM varies depending on the hardware or API, the specific model, and the inference platform employed. In this study, we estimate runtimes for open-source LLMs using vLLM, and for GPT-4o and GPT-4o-mini through the OpenAI API. Except for DeepSeek-R1, we evaluate with the Alibaba Bailian API. The estimated time of our process-based spatiotemporal reasoning evaluation is shown in Table \ref{tab:model_inference}.

\begin{table}[h]
\vspace{-10pt}
\centering
\caption{Confidence Interval of Reasoning Ability Evaluations}
\resizebox{\columnwidth}{!}{
\begin{tabular}{l l l c c c}
\toprule
\textbf{Model} & \textbf{Device or API} & \textbf{Platform} & \textbf{Inference Speed} & \textbf{Batch Size} & \textbf{Total Time} \\
\midrule
Qwen2.5-32B     & 2*A800      & vLLm  & 14.12 s/batch & 32  & 9.9 h \\
GPT-4o          & OpenAI API  & -     & 8.20 s/batch & 16 & 11.50 h  \\
QwQ-32B         & 2*A800      & vLLm  & 30.21 s/batch  & 32 & 18.64 h   \\
DeepSeek-R1     & Alibaba Bailian API & -     & 97.33 s/batch & 32  & 60.05 h \\
o4-mini         & OpenAI API  & -     & 5.61 s/batch   & 16 & 7.87 h    \\
\bottomrule
\end{tabular}}
\label{tab:model_inference}
\vspace{-10pt}
\end{table}

\vspace{-5pt}
\section{Ethics and Societal Impact}\label{sec:social_impact}
\vspace{-5pt}

USTBench aims to systematically evaluate the spatiotemporal reasoning capabilities of LLMs in urban applications such as traffic control, mobility prediction, and urban planning. To ensure privacy, our benchmark exclusively uses publicly available datasets and ensures that no personally identifiable information is included. While these evaluations advance the understanding of LLM capabilities, they also underscore the responsibility to ensure that such models are used ethically. LLM-driven urban agents could influence public infrastructure, mobility patterns, and access to services. Therefore, deploying these models without proper oversight or fail-safes could lead to unintended negative outcomes, especially for vulnerable populations in urban settings.

\vspace{-5pt}
\section{Details of Environments}
\vspace{-5pt}

\subsection{Urban Task Suite and Environment Platform}\label{subsec:urban_tasks}

\textbf{Next POI Prediction}: Predicts movement patterns within urban areas, helping to understand how individuals travel between points of interest (POIs). Following the settings proposed by \citep{wang2023would}, we use FourSquare datasets \citep{foursquare_dataset} for evaluation.

\textbf{Socio-Economic Prediction}: Estimates future socio-economic indicators (\ie GDP) based on historical observations and population density. We collect time-series data on GDP and population development from 2000 to 2019 \cite{wang2022global, worldpop_hub} in Guangzhou, China, for future value prediction.

\textbf{Congestion Prediction}: Anticipates areas where traffic flow will become heavy. We follow the settings of \citep{cheng2018deeptransport} and evaluate LLMs using traffic data from Beijing, China.

\textbf{Traffic-OD Prediction}: Estimates vehicle flow between origin and destination regions. We follow the short-term prediction settings of \cite{yuan2024unist} and use taxi traffic flow datasets from New York, USA.

\textbf{Traffic Signal Control}: Optimizes traffic signal timing to improve traffic flow. Following the settings of \citep{lai2023llmlight, yuan2025collmlight}, we evaluate performance using traffic flow and road network datasets from Hangzhou, China \citep{traffic-signal-control}. The simulation environment is built based on CityFlow \citep{tang2019cityflow}.
    
\textbf{POI Placement}: Determines optimal locations for urban services, such as shops and restaurants, ensuring strategic placement to serve the population and reduce congestion. For evaluation, we follow the settings of \citep{von2022reinforcement} and collect a charging station placement dataset in Guangzhou, China.

\textbf{Road Planning}: Involves designing and optimizing road networks to enhance transportation efficiency. This task requires analyzing road connectivity and infrastructure needs. We follow the settings and environment of \citep{zheng2023road} and use road network data from Cape Town, South Africa, for evaluation.
    
\textbf{Route Planning}: Determines optimal paths for vehicles, considering traffic conditions, distance, and travel time. Following the settings of \citep{li2021towards}, we use New York's road network and simulate traffic using the SUMO traffic simulator \citep{alvarez_lopez_2025}.
    
\textbf{Urban Planning}: Involves designing urban spaces for sustainable, efficient, and livable cities. Following \cite{zheng2023spatial}, we use urban geospatial data from Beijing, China, and build interaction environments.

\subsection{Environment Configuration}\label{subsec:env_config}

For process-based spatiotemporal reasoning evaluation, we generate QA instances using UAgentEnv. Environmental observations in decision-making tasks are collected using a semi-stochastic policy with an exploration coefficient $\epsilon = 0.1$ to ensure diversity. Ground-truth answers for planning QA are derived through a feedback-driven exploratory process, with a planning horizon $H = 5$ and a discount factor $\gamma = 0.9$.
During downstream task evaluation, we configure historical observation and prediction windows based on task characteristics. For socio-economic prediction, we use a 6-year observation window and a 3-year prediction window. For traffic-related tasks—congestion prediction and traffic OD prediction, we adopt a 12-step observation window and a 12-step prediction window. In the next POI prediction task, the agent receives a 30-day activity history and previous visits on the same day and predicts the next visited POI.
For route planning, we synthesize urban mobility patterns using the gravity model \citep{li2021towards}, calibrated on real-world population distributions in New York City. The configurations for decision-making environments follow established benchmarks and experimental protocols from prior work \citep{wei2019colight, von2022reinforcement, li2021towards, zheng2023road, zheng2023spatial}.

\subsection{Agent Prompt}\label{subsec:agent_prompt}

Our agent framework incorporates three prompt templates designed for task-solving, reflection, and memory storage. In the \textit{task-solving prompt}, the agent is instructed to perform a multi-stage reasoning process: 1) it first interprets the spatiotemporal context of the environment; 2) then either forecasts future urban states or predicts the outcomes of candidate actions; 3) finally, the agent outputs anticipated urban states for prediction tasks or selects an optimal action for decision-making tasks.
After execution and receiving environmental feedback, the agent is instructed by the \textit{reflection prompt} to evaluate the effectiveness of its prior decision and summarizes the outcome as an experience to inform future reasoning. These experiences are subsequently aggregated and stored in memory to support continual adaptation with the \textit{summary prompt}. The detailed prompt templates used in each stage are provided in Table \ref{tab:task_prompt}-\ref{tab:summary_prompt}.

\vspace{-5pt}
\section{LLM Post-Training}\label{sec:llm_post_training}
\vspace{-5pt}

\subsection{Instruction Construction}

We post-train Qwen2.5-7B using a synthetic instruction tuning dataset, designed to enhance the model’s capability to interpret spatiotemporal dynamics in urban scenarios. To generate instructions, we first prompt GPT-4o to produce diverse urban scenarios involving various entities and events. Each scenario is designed to elicit a specific spatiotemporal understanding ability, prompting the model to analyze spatial relationships or temporal patterns:

\textbf{Distance}: We randomly assign distances between roads, routes, or urban entities. The model is asked to identify the longest or shortest elements or to compare distances between pairs of entities.

\textbf{Adjacency \& Connectivity}: We Erd\H{o}s--R'enyi (ER) model \citep{erdos1960evolution} to generate random spatial graphs representing urban layouts. The model is then asked to determine adjacency (\ie nearby neighbors) or connectivity (\ie path existence between entities).

\textbf{Duration}: We simulate urban events (\eg travel, wait times) with randomly assigned durations. The model is tasked with identifying the longest or shortest event or comparing durations between events.

\textbf{Chronology}: We use POI check-in data from Tokyo \citep{foursquare_dataset} (distinct from the dataset used in USTBench). The model is tasked to identify the correct temporal sequence of check-in events.

\textbf{Trend, Local Extrema, and Periodicity}: We leverage real-world urban time-series datasets, including PEMS04 (traffic flow), Solar (solar power output), and Electricity (power usage). The model is instructed to identify global trends (\eg increasing/decreasing), local extrema (\eg peak hours), and periodicity (\eg daily/weekly cycles).

\subsection{Supervised Distillation Fine-Tuning}

Leveraging the constructed instructions described above, we collect responses from DeepSeek-R1 using rejection sampling to ensure high-quality outputs. Representative examples of the instruction tuning data are provided in Table \ref{tab:distance_prompt}-\ref{tab:periodicity_prompt}. For supervised fine-tuning, we adopt Llama-Factory \citep{zheng2024llamafactory} with Low-Rank Adaptation (LoRA) for training. The learning rate is set to $1 \times 10^{-4}$.

\vspace{-5pt}
\section{Detailed Evaluations}\label{sec:detailed_eval}
\vspace{-5pt}

\subsection{Downstream Task Evaluation Metrics}
We evaluate LLMs' downstream task performance across nine urban tasks using task-specific metrics. \textit{Socio-economic prediction}: We apply Mean Absolute Percentage Error (MAPE) to assess GDP forecasting accuracy over a three-year window.  \textit{Congestion prediction}: This task is framed as a five-level classification task (levels 0–4), and is evaluated using MAPE and accuracy. \textit{Traffic-OD prediction}: We adopt Mean Absolute Error (MAE) and Symmetric Mean Absolute Percentage Error (SMAPE) to evaluate forecasting accuracy for vehicle arrivals and departures. \textit{Next POI prediction}: We use Precision and Mean Reciprocal Rank (MRR) to measure the recall quality of predictions. 
\textit{Urban planning}: This task is assessed through accessibility to service facilities and ecological coverage. \textit{Road planning}: The performance is measured by construction costs and the average travel distance to neighboring regions. \textit{POI placement}: The task is evaluated by the average travel and waiting time for urban services. \textit{Traffic signal control}: The performance is measured using average travel time (ATT) and waiting time (AWT) in the road network. \textit{Route planning}: The task is evaluated using average travel time and network throughput.

\vspace{-5pt}
\subsection{Baseline Configuration}\label{subsec:baselines}

\textbf{LLM Configuration}: In this study, we mainly evaluate open-source LLMs using the vLLM inference framework. For proprietary models such as GPT-4o and GPT-4o-mini, we utilize the OpenAI API, while DeepSeek-R1 is evaluated via the Alibaba Bailian API. All evaluations are conducted with a fixed decoding temperature of 0.1 to ensure reproducibility. Inference of open-sourced LLMs is performed on a server equipped with two NVIDIA A800-80GB GPUs.

\textbf{Domain-specific Method Configuration}: To provide a comprehensive performance comparison, we also benchmark LLMs against traditional methods widely used in each domain: For time-series forecasting tasks (urban development prediction and congestion detection), we use ARIMA \citep{box2013box} as the baseline model. 
For urban planning, we employ a geometric set coverage algorithm (GSCA) \citep{wei2016coverage}, which solves a geometric set-coverage-like problem by maximizing the spatial coverage of designated land-use types.
For road planning, we apply a genetic algorithm \citep{gad2024pygad}, where a linear layer represents road features, and roads are incrementally constructed based on learned sampling probabilities.

\vspace{-5pt}
\subsection{Downstream Task Performance}

The downstream task performance of representative LLMs is shown in Table \ref{tab:all_task_evaluation}. Notably, reasoning LLMs do not consistently outperform their non-reasoning base models in real-world urban scenarios. This indicates that advances in general mathematical and logical reasoning do not necessarily benefit urban tasks. The finding underscores the importance of developing domain-specific approaches tailored to the unique challenges of urban spatiotemporal reasoning.

\begin{table}[h]
\vspace{-10pt}
\renewcommand{\arraystretch}{1.1}
\setlength{\tabcolsep}{1pt}
\centering
\caption{Performance on downstream urban tasks. Lower values ($\downarrow$) for MAPE, SMAPE, MAE, Cost, Distance, ATT, and AWT indicate better performance. Higher values ($\uparrow$) for Accuracy, Service, Ecology, Precision, MRR, and Throughput indicate better outcomes.}
\resizebox{\textwidth}{!}{
\begin{tabular}{cc|cc|cc|cc|cc|cc|cc|cc|cc}
\toprule
\multirow{3}{*}{\textbf{Model}} & \textbf{Socio-economy} & \multicolumn{2}{c|}{\textbf{Congestion}} & \multicolumn{2}{c|}{\textbf{Urban}} & \multicolumn{2}{c|}{\textbf{Road}} & \multicolumn{2}{c|}{\textbf{Traffic-OD}} & \multicolumn{2}{c|}{\textbf{Next POI}} & \multicolumn{2}{c|}{\textbf{POI}} & \multicolumn{2}{c|}{\textbf{Signal}} & \multicolumn{2}{c}{\textbf{Route}} \\
& \textbf{Prediction} & \multicolumn{2}{c|}{\textbf{Prediction}} & \multicolumn{2}{c|}{\textbf{Planning}} & \multicolumn{2}{c|}{\textbf{Planning}} & \multicolumn{2}{c|}{\textbf{Prediction}} & \multicolumn{2}{c|}{\textbf{Prediction}} & \multicolumn{2}{c|}{\textbf{Placement}} & \multicolumn{2}{c|}{\textbf{Control}} & \multicolumn{2}{c}{\textbf{Planning}} \\
\cline{2-18}
& MAPE (\%) & MAPE (\%) & Acc (\%) & Serv & Eco & Cost & Dist & MAE & SMAPE (\%) & Prec@10 & MRR@10 & ATT & AWT & ATT & AWT & ATT & Thruput \\
\hline
\multicolumn{18}{c}{\textbf{Non-Reasoning LLMs}} \\
\hline
Qwen2.5-7B & 34.57 & 66.19 & 40.51 & 0.5951 & \textbf{0.6440} & 20.72 & 1.50 & \textbf{5.09} & \textbf{13.35} & 0.3787 & 0.1888  & 1.21 & 0.56 & \textbf{820.28} & \textbf{472.20} & 1417.93 & 367 \\
GLM4-9B & 58.43 & 41.41 & 54.71 & 0.6355 & 0.4507 & 20.59 & 1.50 & 84.22 & 96.85 & \textbf{0.7017} & 0.4405 & 1.17 & \underline{0.25} & 1109.80 & 626.89 & 1366.45 & 372 \\
Qwen2.5-32B & 6.00 & \textbf{24.90} & \textbf{65.90} & 0.6335 & 0.5209 & 20.56 & 1.55 & \underline{8.11} & 33.43 & \underline{0.6627} & 0.5096  & 1.16 & 0.47 & 1189.31 & 672.84 & 1376.33 & 376 \\
GLM4-32B & 9.41 & 28.61 & \underline{63.02} & 0.6662 & 0.4715 & \underline{18.44} & 1.52 & 11.20 & 43.70 & 0.5183 & 0.3786 & 1.12 & 0.36 & 1290.61 & 690.64 & 1384.54 & \textbf{373} \\
Llama3.3-70B & 10.86 & 38.88 & 56.10 & \underline{0.6561} & \underline{0.5842} & 19.10 & 1.57 & 8.52 & \underline{33.13} & 0.4863 & 0.3273 & 1.12 & 0.36 & 1324.84 & 682.10 & \underline{1310.11} & 370 \\
\hline
\multicolumn{18}{c}{\textbf{Reasoning LLMs}} \\
\hline
DeepSeek-R1- & \multirow{2}{*}{79.23} & \multirow{2}{*}{67.42} & \multirow{2}{*}{37.88} & \multirow{2}{*}{0.6348} & \multirow{2}{*}{0.6111} & \multirow{2}{*}{20.60} & \multirow{2}{*}{1.47} & \multirow{2}{*}{140.05} & \multirow{2}{*}{130.84} & \multirow{2}{*}{0.2910} & \multirow{2}{*}{0.1455} & \multirow{2}{*}{1.18} & \multirow{2}{*}{0.46} & \multirow{2}{*}{1000.56} & \multirow{2}{*}{\underline{541.84}} & \multirow{2}{*}{1390.65} & \multirow{2}{*}{372} \\
Distill-Qwen-7B & & & & & & & & & & & & & & & & & \\
\hline
GLM-Z1-9B & 11.58 & 45.87 & 52.01 & 0.6443 & 0.5430 & 18.80 & \underline{1.33} & 70.03 & 92.38 & 0.5637 & 0.4963  & 1.17 & 0.52 & \underline{970.26} & 711.33 & 1283.29 & 371 \\
QwQ-32B & \textbf{5.64} & 44.89 & 52.88 & \textbf{0.6751} & 0.5792 & \textbf{18.40} & 1.77 & 8.16 & 36.96 & \underline{0.6817} & \textbf{0.6013}  & 1.19 & 0.54 & 1267.82 & 672.88 & 1417.32 & 373 \\
GLM-Z1-32B & 7.55 & 47.93 & 51.22 & 0.6468 & 0.3965 & 18.57 & 1.87 & 56.46 & 77.46 & 0.6430 & \underline{0.5540}  & \textbf{1.10} & 0.74 & 1132.48 & 641.73 & 1331.51 & 370 \\
\hline
DeepSeek-R1- & \multirow{2}{*}{\underline{5.94}} & \multirow{2}{*}{\underline{38.78}} & \multirow{2}{*}{55.50} & \multirow{2}{*}{0.6560} & \multirow{2}{*}{0.4711} & \multirow{2}{*}{19.42} & \multirow{2}{*}{\textbf{1.13}} & \multirow{2}{*}{17.79} & \multirow{2}{*}{45.23} & \multirow{2}{*}{0.5530} & \multirow{2}{*}{0.4732} & \multirow{2}{*}{\underline{1.12}} & \multirow{2}{*}{\textbf{0.24}} & \multirow{2}{*}{1202.11} & \multirow{2}{*}{629.29} & \multirow{2}{*}{\textbf{1267.20}} & \multirow{2}{*}{\underline{375}} \\
Distill-Llama-70B & & & & & & & & & & & & & & & & & \\
\bottomrule
\end{tabular}}
\label{tab:all_task_evaluation}
\vspace{-10pt}
\end{table}

\subsection{Reasoning Behavior Analysis}

\subsubsection{Cost-Effectiveness Analysis}

\begin{wrapfigure}{r}{0.5\textwidth}
    \vspace{-15pt}
    \centering
    \includegraphics[width=0.48\textwidth]{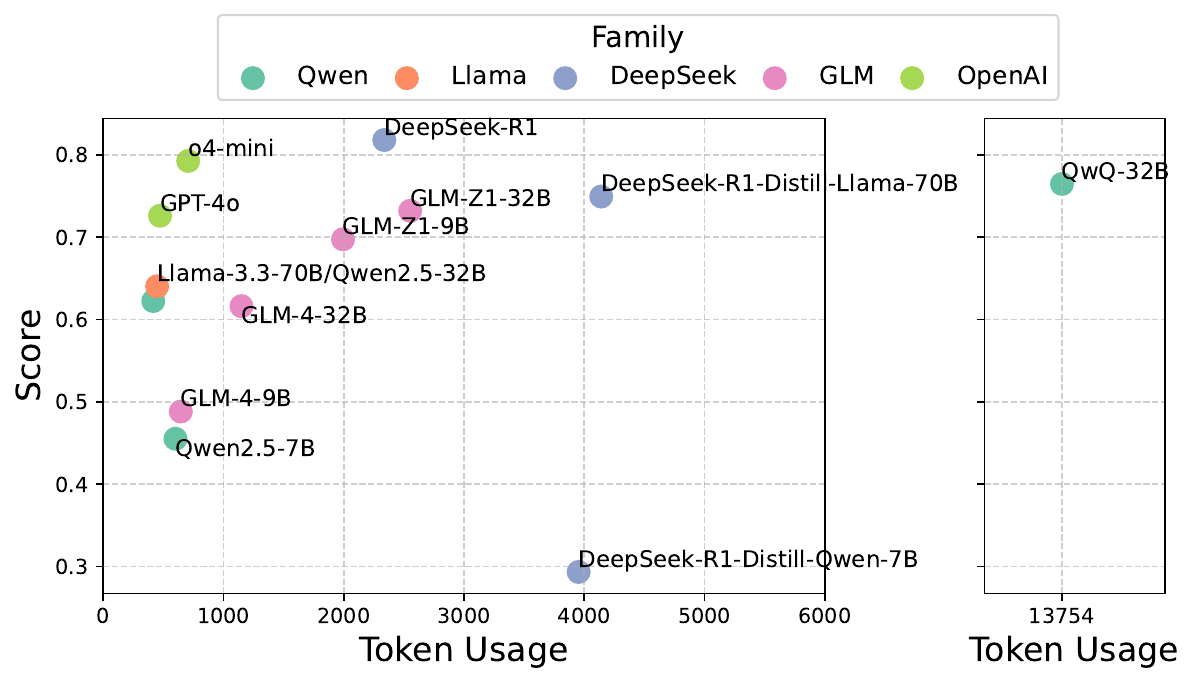}
    \caption{The token usage vs. score.}
    \label{fig:token_vs_score}
    \vspace{-10pt}
\end{wrapfigure}

In our process-based spatiotemporal reasoning evaluation, we observe that the non-reasoning model GPT-4o is comparable, and in some cases surpasses, reasoning models. To further explore this, we conduct a cost-effectiveness analysis (Figure \ref{fig:token_vs_score}), comparing model performance on spatiotemporal understanding relative to the number of reasoning tokens used. Among all models, o4-mini demonstrates the highest cost-efficiency, achieving strong performance with minimal reasoning overhead, followed closely by GPT-4o. In contrast, while DeepSeek-R1 delivers strong performance, its reasoning processes are often verbose and time-consuming, making it less suitable for real-time deployment scenarios (\eg traffic management). These findings highlight research opportunities to develop lightweight and efficient paradigms for urban spatiotemporal reasoning, drawing inspiration from designs like o4-mini and GPT-4o.

\subsubsection{Repetition Issues}\label{subsubsec:repetition}
\begin{figure}[h]
\centering
\includegraphics[width=\textwidth]{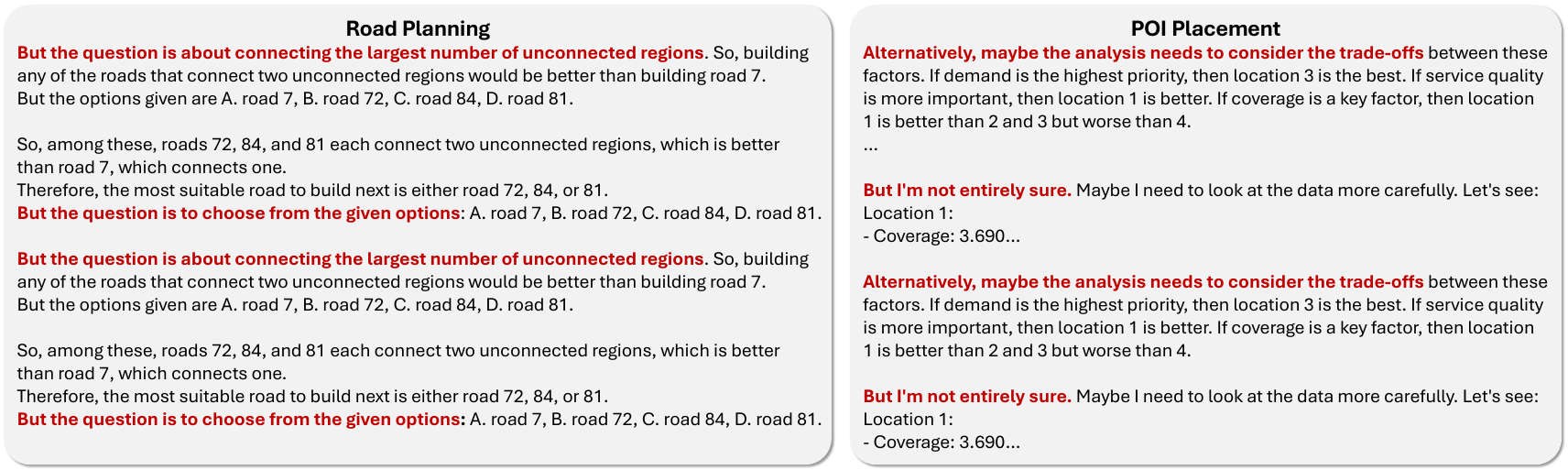}
\centering
\caption{The repetition issues of DeepSeek-R1-Distill-Qwen-7B.}
\label{fig:repetition}
\vspace{-10pt}
\end{figure}

We observe severe repetition issues on DeepSeek-R1-Distill-Qwen2.5-7B, where it underperforms its base model, Qwen2.5-7B, on several tasks. Upon analyzing its reasoning traces, we find that the model tends to repeat certain thought patterns, called "aha moments" \citep{guo2025deepseek}, without progressing toward a solution. This repetitive behavior resembles ineffective in-context learning loops, and appears more pronounced in specialized domains like urban tasks. While such issues have been partially mitigated in mathematical and logical reasoning, they remain a persistent challenge in other specialized domains. Figure \ref{fig:repetition} provides examples of failed reasoning processes.

\subsubsection{Process-based Spatiotemporal Reasoning QA Examples} \label{subsubsec:prompt_response}

\begin{table*}[hb]
  \centering
  \caption{The spatiotemporal understanding QA for distance relation analysis.}
  \begin{tabularx}{\textwidth}{X}
    \toprule
    \textbf{Task Description} \\[6pt]
    We are tasked with determining the optimal location for a new electric vehicle (EV) charging station to maximize benefits by effectively meeting demand. The candidate locations, along with their expected charging station coverage, demand (number of vehicles), average distance, travel time, and waiting time to existing charging stations, are provided below. \\[12pt]
    \\
    \textbf{Data Schema} \\[6pt]
    - coverage: The number of locations that will be within the influence radius of the charging station after its construction.\\
    - demand: The number of vehicles requiring charging near the location.\\
    - distance: The average distance that vehicles near the location must travel to reach existing charging stations.\\
    - waiting\_time: The average waiting time at nearby existing charging stations. NaN indicates no station nearby and the highest waiting time.\\
    - charging\_time: The average charging time at nearby existing charging stations. NaN indicates no station nearby and the highest charging time.\\[12pt]
    \\
    \textbf{Data} \\[6pt]
    location 1: \\
    - coverage: 3.315 \\
    - demand: 43531.73 \\
    - distance: 265.80m \\
    - waiting\_time: 0.00h \\
    - charging\_time: 1.80h \\[6pt]
    location 2: \\
    - coverage: 3.316 \\
    - demand: 74504.47 \\
    - distance: 159.25m \\
    - waiting\_time: 0.04h \\
    - charging\_time: 20.61h \\[6pt]
    location 3: \\
    - coverage: 3.316 \\
    - demand: 130231.14 \\
    - distance: 160.19m \\
    - waiting\_time: 0.04h \\
    - charging\_time: 21.75h \\[6pt]
    location 4: \\
    ...\\
    \\
    \textbf{Question} \\[6pt]
    Based on the data provided, rank the locations by average distance to nearby charging stations, from the farthest to the closest.\\[6pt]
    A. 1 > 2 > 3 > 4 \\
    B. 2 > 4 > 3 > 1 \\
    C. 1 > 2 > 3 > 4 \\
    D. 4 > 1 > 3 > 2 \\[12pt]
    
    \textbf{Note} \\[6pt]
    Please answer the question using the information given. Let’s work this out in a step-by-step way to be sure that we have the right answer. Then provide your final answer within the XML tag \texttt{<Answer>A/B/C/D</Answer>}. \\
    \bottomrule
  \end{tabularx}
  \vspace{-80pt}
  \label{tab:st_under_dist_q}
\end{table*}

\begin{figure}[h]
\centering
\includegraphics[width=\textwidth]{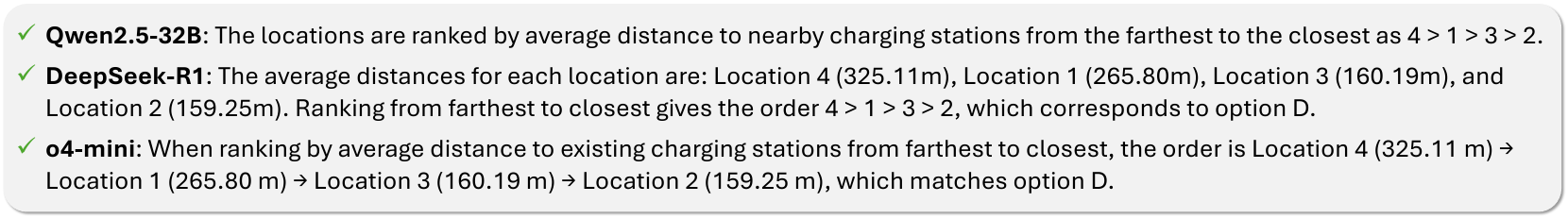}
\centering
\caption{The LLMs' reasoning on distance relation analysis.}
\label{fig:st_under_dist_a}
\vspace{-5pt}
\end{figure}

\begin{table*}[h]
\vspace{-5pt}
  \centering
  \caption{The spatiotemporal understanding QA for adjacency relation analysis.}
  \vspace{-5pt}
  \begin{tabularx}{\textwidth}{X}
    \toprule
    \textbf{Task Description} \\[6pt]
    You are tasked with predicting urban indicators—specifically, the GDP and population of Guangzhou for the current year. Historical data for the GDP and population of the target region and its nearby regions will be provided. Additionally, an example region with its GDP and population records for previous years and the current year will be included for reference. \\[12pt]
    \textbf{Data Schema} \\[6pt]
    - gdp\_in\_previous\_years: The GDP (in millions of USD) of a region in past years. \\
    - population\_in\_previous\_years: The population of a region in past years. \\
    - connectivity: A list of tuples in the format (region\_1, region\_2, distance), defining undirected edges between regions with the specified distance in m. \\[12pt]
    
    \textbf{Data} \\[6pt]
    \textit{Target Region 442:} \\
    - gdp\_in\_previous\_years: [(8.54, 2000), ..., (23.19, 2018)] \\
    - population\_in\_previous\_years: [(2942.97, 2000), ..., (2242.3, 2018)] \\[6pt]

    \textit{Nearby Region 418:} \\
    - gdp\_in\_previous\_years: [(22.55, 2000), ..., (55.37, 2018)] \\
    - population\_in\_previous\_years: [(2589.67, 2000), ..., (8729.58, 2018)] \\[6pt]

    ...

    \textit{Adjacency Info:} \\
    \texttt{[(442, 418, 7.58km), (442, 443, 6.09km), (442, 467, 4.51km), ..., (467, 492, 4.51km)]} \\[6pt]

    \textit{Example Region 192:} \\
    - gdp\_records: [(201.59, 2000), ..., (628.55, 2019)] \\
    - population\_records: [(5908.03, 2000), ..., (11102.63, 2019)] \\
    
    \textbf{Question} \\[6pt]
    Based on the connectivity data, rank spatial correlation from the target region to its nearby regions, ordering them from the strongest to the weakest correlation. \\[6pt]
    A. 492 > 468 > 417 > 443. \\
    B. 443 > 492 > 468 > 417. \\
    C. 492 > 468 > 417 > 443. \\
    D. 468 > 443 > 492 > 417. \\[6pt]

    \textbf{Note} \\[6pt]
    Please answer the question using the information given. Let’s work this out in a step-by-step way to be sure that we have the right answer. Then provide your final answer within the XML tag \texttt{<Answer>A/B/C/D</Answer>}. \\
    \bottomrule
  \end{tabularx}
  \vspace{-10pt}
  \label{tab:st_under_adj_q}
\end{table*}

\begin{figure}[h]
\centering
\includegraphics[width=\textwidth]{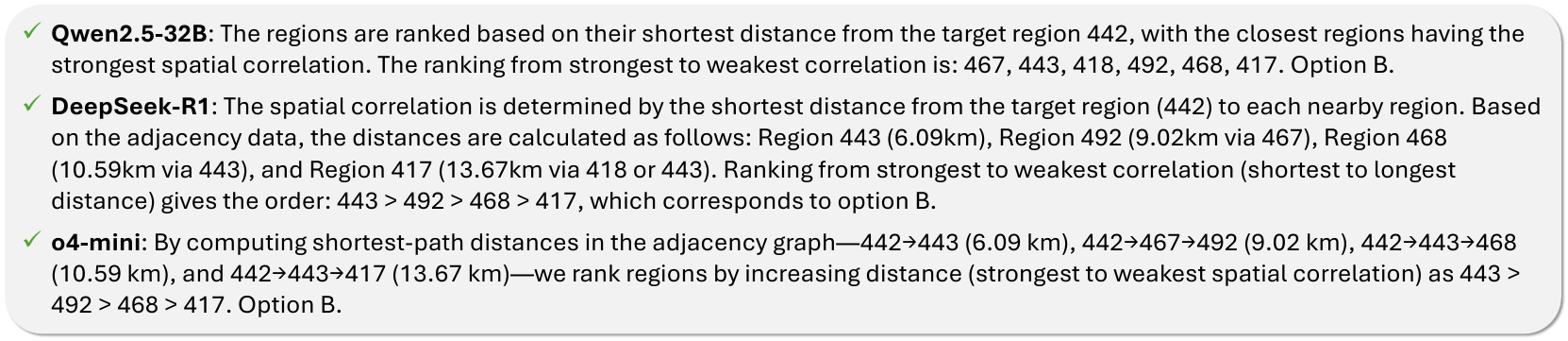}
\centering
\caption{The LLMs' reasoning on adjacency relation analysis.}
\label{fig:st_under_adj_a}
\vspace{-5pt}
\end{figure}

\begin{table*}[h]
  \vspace{-5pt}
  \centering
  \caption{The spatiotemporal understanding QA for connectivity relation analysis.}
  \begin{tabularx}{\textwidth}{X}
    \toprule
    \textbf{Task Description} \\[6pt]
    You are a traffic signal control agent at a target intersection, managing traffic from four directions: north, south, east, and west. The intersection has 12 lanes, categorized as: NL, NT, NR, SL, ST, SR, EL, ET, ER, WL, WT, and WR. Vehicles at the intersection travel at an average speed of 11 meters per second. The real-time traffic conditions for each lane are provided below. If a lane's traffic condition is not specified, it indicates that no vehicles are currently present in that lane. \\
    \\
    \textbf{Data Schema} \\[6pt]
    - queue: Number of vehicles waiting at the lane for a green signal.\\
    - move: Number of vehicles currently moving through the lane.\\
    - wait\_time: Average wait time (in minutes) for vehicles in the queue.\\
    - occupancy: Lane occupancy rate (0 = free-flow, 100\% = fully congested).\\
    - connectivity: A list of tuples in the format (lane\_1, lane\_2, distance), defining a directed connection between lanes with the specified distance in m. The traffic flow from lane\_1 (upstream) will arrive at lane\_2 (downstream) in the future. \\
    \\
    \textbf{Data} \\[6pt]
    Target Intersection 112:\\
    - SL lane:\\
    \quad - queue: 37.0\\
    \quad - move: 0\\
    \quad - wait\_time: 2.86\\
    \quad - occupancy: 112.12\%\\
    - ET lane:\\
    \quad - queue: 18.0\\
    \quad - move: 0\\
    \quad - wait\_time: 4.26\\
    \quad - occupancy: 54.55\%\\
    ...\\
    \\
    Neighboring Intersection 140:\\
    - ST lane:\\
    \quad - queue: 4.0\\
    \quad - move: 0\\
    \quad - wait\_time: 0.45\\
    \quad - occupancy: 12.12\%\\
    - SL lane:\\
    \quad - queue: 3.0\\
    \quad - move: 1\\
    \quad - wait\_time: 0.41\\
    \quad - occupancy: 12.12\%\\
    ...
    \\
    Other Neighbors... \\
    \\
    The connectivity of lanes:\\
    \texttt{[(NT of 113, NT of 112, 300m), (EL of 113, NT of 112, 300m), (WR of 113, NT of 112, 300m), (NT of 113, NL of 112, 300m), ..., (SR of 84, WR of 112, 300m)]} \\
    \\
    \textbf{Question} \\[6pt]
    Based on the data provided, select the correct ranking of neighboring upstream lanes of the target intersection by the moving vehicles, ordering them from the highest to the lowest.\\
    A. ST of 140 > EL of 140 = EL of 112 > other lanes\\
    B. WR of 113 > ET of 84 = ET of 112 > other lanes\\
    C. ET of 140 = ST of 140 > ET of 84 > other lanes\\
    D. WR of 113 > ET of 140 = SL of 140 > other lanes \\
  \end{tabularx}
  \vspace{-10pt}
  \label{tab:st_under_con_q}
\end{table*}

\begin{table*}[t]
  \centering
  \vspace{-5pt}
  \begin{tabularx}{\textwidth}{X}
    \textbf{Note} \\[6pt]
    Please answer the question using the information given. Let's work this out in a step-by-step way to be sure that we have the right answer. Then provide your final answer within the XML tag \texttt{<Answer>A/B/C/D</Answer>}.
    \\
    \bottomrule
  \end{tabularx}
  \vspace{-10pt}
\end{table*}

\begin{figure}[t]
\centering
\includegraphics[width=\textwidth]{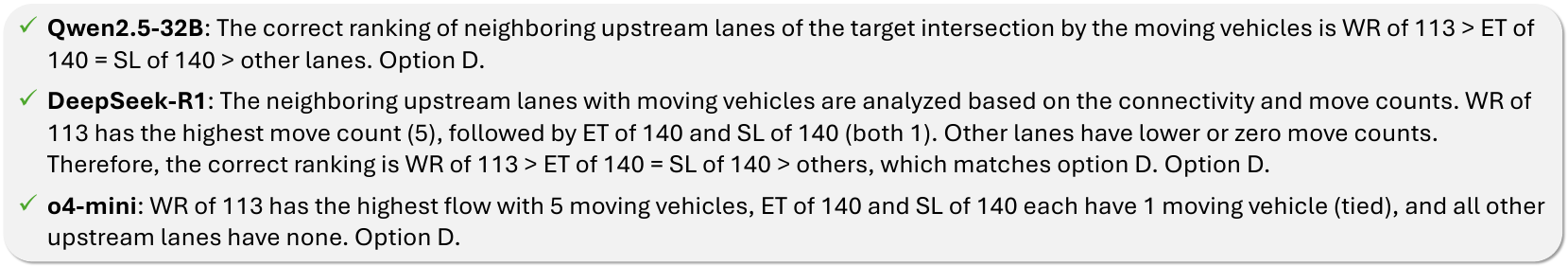}
\centering
\caption{The LLMs' reasoning on connectivity relation analysis.}
\label{fig:st_under_con_a}
\vspace{-5pt}
\end{figure}

\begin{table*}[h]
\vspace{-5pt}
  \centering
  \caption{The spatiotemporal understanding QA for duration pattern analysis.}
  \vspace{-5pt}
  \begin{tabularx}{\textwidth}{X}
    \toprule
    \textbf{Task Description} \\[6pt]
    We are tasked with determining the optimal location for a new electric vehicle (EV) charging station to maximize benefits by effectively meeting demand. The candidate locations, along with their expected charging station coverage, demand (number of vehicles), average distance, travel time, and waiting time to existing charging stations, are provided below. \\
    
    \textbf{Data Schema} \\[6pt]
    - coverage: The number of locations that will be within the influence radius of the charging station after its construction. \\
    - demand: The number of vehicles requiring charging near the location. \\
    - distance: The average distance that vehicles near the location must travel to reach existing charging stations. \\
    - waiting\_time: The average waiting time at nearby existing charging stations. NaN indicates no station nearby and the highest waiting time. \\
    - charging\_time: The average charging time at nearby existing charging stations. NaN indicates no station nearby and the highest charging time. \\

    \textbf{Data} \\[6pt]
    location 1: \\
    \quad - coverage: 3.068 \\
    \quad - demand: 90827.49 \\
    \quad - distance: 188.79m \\
    \quad - waiting\_time: 0.05h \\
    \quad - charging\_time: 22.14h \\
    
    location 2: \\
    \quad - coverage: 3.059 \\
    \quad - demand: 105897.62 \\
    \quad - distance: 197.29m \\
    \quad - waiting\_time: 0.03h \\
    \quad - charging\_time: 13.78h \\
    
    location 3: \\
    \quad - coverage: 3.059 \\
    \quad - demand: 12443.36 \\
    \quad - distance: 371.80m \\
    \quad - waiting\_time: 0.00h \\
    \quad - charging\_time: 0.96h \\
    
    location 4: \\
    \quad - coverage: 3.062 \\
    \quad - demand: 22993.54 \\
    \quad - distance: 304.35m \\
    \quad - waiting\_time: 0.00h \\
    \quad - charging\_time: 1.01h \\
    
    \textbf{Question} \\[6pt]
    Based on the data provided, rank the locations by average charging time at near charging stations, from the longest to the shortest. \\
  \end{tabularx}
  \vspace{-10pt}
  \label{tab:st_under_adj_q}
\end{table*}

\begin{table*}[h]
\vspace{-5pt}
  \centering
  \begin{tabularx}{\textwidth}{X}
    A. 4 > 1 > 2 > 3 \\
    B. 3 > 4 > 1 > 2 \\
    C. 4 > 1 > 2 > 3 \\
    D. 1 > 2 > 4 > 3 \\

    \textbf{Note} \\[6pt]
    Please answer the question using the information given. Let’s work this out in a step-by-step way to be sure that we have the right answer. Then provide your final answer within the XML tag \texttt{<Answer>A/B/C/D</Answer>}. \\
    \bottomrule
  \end{tabularx}
  \vspace{-10pt}
  \label{tab:st_under_dura_q}
\end{table*}

\begin{figure}[t]
\centering
\includegraphics[width=\textwidth]{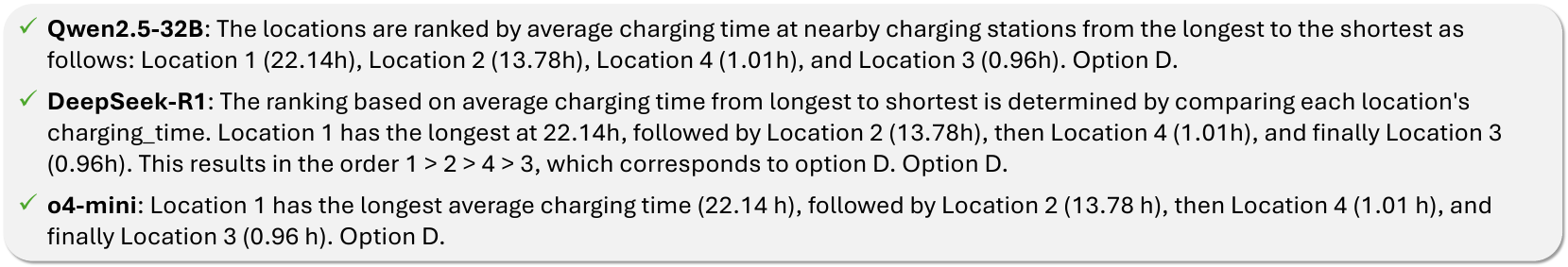}
\centering
\caption{The LLMs' reasoning on duration pattern analysis.}
\label{fig:st_under_dura_a}
\vspace{-5pt}
\end{figure}

\begin{table*}[h]
\vspace{-5pt}
  \centering
  \caption{The spatiotemporal understanding QA for chronological pattern analysis.}
  \vspace{-5pt}
  \begin{tabularx}{\textwidth}{X}
    \toprule
    \textbf{Task Description} \\[6pt]
    You will analyze a user’s historical stay records and predict their visiting patterns. \\

    \textbf{Data Schema} \\[6pt]
    - History: The user's historical travel sequence over the past 30 days. \\
    - Recent: The user's recent visit sequence. \\
    - trajectory: A list of tuples (POI, weekday, time), representing user visits. \\
    - adjacency: A list of tuples (POI\_1, POI\_2, distance), representing spatial relationships between POIs. \\

    \textbf{Data} \\[6pt]
    History: \\[3pt]
    Day 1: \\
    - trajectory: [(Train Station 152, Thursday, 08:10 PM)] \\
    - adjacency: N/A \\[3pt]
    Day 2: \\
    - trajectory: [(Train Station 152, Friday, 10:02 AM), (Building 20014, Friday, 06:01 PM)] \\
    - adjacency: N/A \\[3pt]
    Day 3: \\
    - trajectory: [(Train Station 152, Monday, 10:01 AM), (Building 20014, Monday, 10:13 AM), (Coffee Shop 14357, Monday, 06:37 PM)] \\
    - adjacency: \texttt{[(Building 20014, Train Station 152, 773.0m), (Building 20014, Coffee Shop 14357, 2886.0m)]} \\[3pt]
    ...\\
    Day 30: \\
    - trajectory: [(Train Station 152, Monday, 09:58 AM)] \\
    - adjacency: N/A \\

    \textbf{Question} \\[6pt]
    Based the user’s historical stays, identify the most likely pattern in their behaviors: \\[3pt]
    A. The user usually goes to the Coffee Shop after Building. \\
    B. The user usually goes to the Building after Building. \\
    C. The user usually goes to the Train Station after Train Station. \\
    D. The user usually goes to the Train Station after Building. \\

    \textbf{Note} \\[6pt]
    Please answer the question using the information given. Let’s work this out in a step-by-step way to be sure that we have the right answer. Then provide your final answer within the XML tag \texttt{<Answer>A/B/C/D</Answer>}. \\
    \bottomrule
  \end{tabularx}
  \vspace{-10pt}
  \label{tab:st_under_chron_q}
\end{table*}

\begin{figure}[t]
\centering
\includegraphics[width=\textwidth]{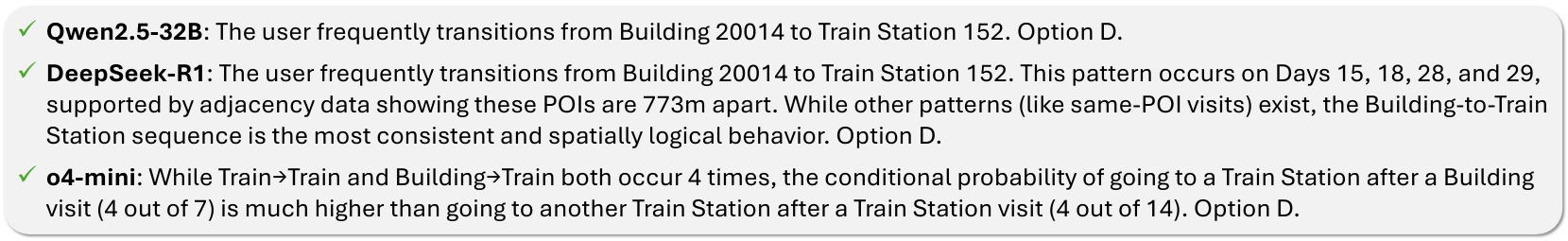}
\centering
\caption{The LLMs' reasoning on chronological pattern analysis.}
\label{fig:st_under_chron_a}
\vspace{-5pt}
\end{figure}

\begin{table*}[h]
\vspace{-5pt}
\centering
\caption{The spatiotemporal understanding QA for trend pattern analysis.}
\vspace{-5pt}
\begin{tabularx}{\textwidth}{X}
\toprule
\textbf{Task Description} \\[6pt]
You are analyzing traffic congestion in a road network of Beijing, China. Given a target road and its nearby neighboring roads, along with time-stamped traffic conditions for each, perform a congestion analysis. \\[12pt]
\\
\textbf{Data Schema} \\[6pt]
- \textit{Target road}: The road for which congestion patterns need to be analyzed.\\
- \textit{Nearby roads}: Roads within up to two hops from the target road.\\
- \textit{speed\_limit}: Speed limit on the road, with levels from 1 (lowest) to 4 (highest).\\
- \textit{congestion\_level}: A series of congestion levels (from 0 to 4, where 4 is the highest) recorded at specific times during the day.\\
- \textit{congestion\_level\_in\_past\_3\_days}: Historical time-series data of congestion levels over the past three days, recorded at specific time intervals.\\
- \textit{Connectivity}: A list of tuples in the format (road\_1, road\_2, distance), defining directed connections between roads with the specified distance in meters. \\[12pt]

\textbf{Data} \\[6pt]
\textit{Target road}:\\
road 33574:\\
- speed\_limit: \\
- congestion\_level: [(4, '17:00'), (4, '17:05'), (4, '17:10'), (4, '17:15'), (4, '17:20'), (4, '17:25'), (4, '17:30'), (3, '17:35'), (3, '17:40'), (3, '17:45'), (2, '17:50'), (3, '17:55')]\\
- congestion\_level\_in\_past\_3\_days: [('N/A', '00:00'), ('N/A', '01:00'), ..., (1, '22:00'), (1, '23:00')]\\

\textit{Nearby roads}:

road 27810:\\
- speed\_limit: 3\\
- congestion\_level: [(1, '17:00'), (1, '17:05'), (1, '17:10'), (1, '17:15'), (1, '17:20'), (1, '17:25'), (1, '17:30'), (1, '17:35'), (1, '17:40'), (1, '17:45'), (1, '17:50'), (1, '17:55')]\\
... \\

\textit{Connectivity of roads}:
[(22598, 22901, 500m), (27810, 27809, 500m), (33574, 27810, 500m), (33574, 28036, 500m), (28036, 28035, 500m), (22901, 33574, 500m)] \\[12pt]
\\
\textbf{Question} \\[6pt]
Analyze the traffic conditions for target road 33574. What is the overall trend in traffic congestion?\\
A. No trend \\
B. Increase \\
C. Decrease \\
D. No trend first, then increase \\
E. No trend first, then decrease \\
F. Increase first, then no trend \\
G. Decrease first, then no trend \\
H. Increase first, then decrease \\
I. Decrease first, then increase \\

\textbf{Note} \\[6pt]
Please answer the question using the information given. Let’s work this out in a step-by-step way to be sure that we have the right answer. Then provide your final answer within the XML tag \texttt{<Answer>A/B/C/D/E/F/G/H/I</Answer>}. \\
\bottomrule
\end{tabularx}
\vspace{-10pt}
\label{tab:st_under_trend_q}
\end{table*}

\begin{figure}[t]
\centering
\includegraphics[width=\textwidth]{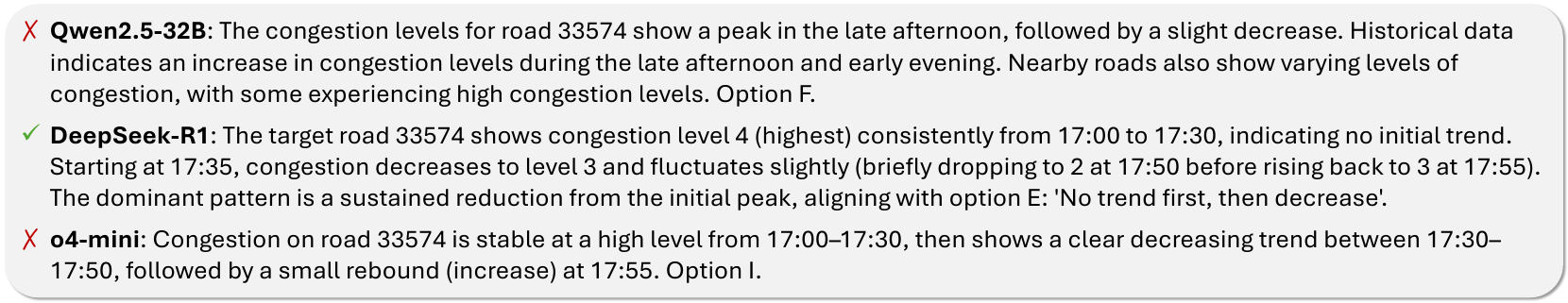}
\centering
\caption{The LLMs' reasoning on trend pattern analysis.}
\label{fig:st_under_trend_a}
\vspace{-5pt}
\end{figure}

\begin{table*}[h]
\vspace{-5pt}
  \centering
  \caption{The spatiotemporal understanding QA for local extrema analysis.}
  \vspace{-5pt}
  \begin{tabularx}{\textwidth}{X}
    \toprule
    \textbf{Task Description} \\[6pt]
    You are analyzing traffic congestion in a road network of Beijing, China. Given a target road and its nearby neighboring roads, along with time-stamped traffic conditions for each, perform a congestion analysis. \\
    
    \textbf{Data Schema} \\[6pt]
    - Target road: The road for which congestion patterns need to be analyzed. \\
    - Nearby roads: Roads within up to two hops from the target road. \\
    - speed\_limit: Speed limit on the road, with levels from 1 (lowest) to 4 (highest). \\
    - congestion\_level: A series of congestion levels (from 0 to 4, where 4 is the highest) recorded at specific times during the day. \\
    - congestion\_level\_in\_past\_3\_days: Historical time-series data of congestion levels over the past three days, recorded at specific time intervals. \\
    - Connectivity: A list of tuples in the format (road\_1, road\_2, distance), defining directed connections between roads with the specified distance in meters. \\
    
    \textbf{Data} \\[6pt]
    Target road: \\
    road 27303: \\
    - speed\_limit: 4 \\
    - congestion\_level: [(1, '06:00'), (1, '06:05'), (1, '06:10'), (1, '06:15'), (1, '06:20'), (1, '06:25'), (1, '06:30'), (1, '06:35'), (2, '06:40'), (2, '06:45'), (1, '06:50'), (2, '06:55')] \\
    - congestion\_level\_in\_past\_3\_days: [(1, '00:00'), (1, '01:00'), (1, '02:00'), (1, '03:00'), (1, '04:00'), (1, '05:00'), (1, '06:00'), (1, '07:00'), (1, '08:00'), (1, '09:00'), (1, '10:00'), (1, '11:00'), (1, '12:00'), (1, '13:00'), (1, '14:00'), (1, '15:00'), (1, '16:00'), (1, '17:00'), (1, '18:00'), (2, '19:00'), (2, '20:00'), (1, '21:00'), (1, '22:00'), (1, '23:00'), (1, '00:00'), (1, '01:00'), (1, '02:00'), (1, '03:00'), (1, '04:00'), (2, '05:00'), (1, '06:00'), (2, '07:00'), (1, '08:00'), (1, '09:00'), (1, '10:00'), (1, '11:00'), (1, '12:00'), (1, '13:00'), (1, '14:00'), (1, '15:00'), (1, '16:00'), (1, '17:00'), (1, '18:00'), (2, '19:00'), (1, '20:00'), (1, '21:00'), (1, '22:00'), (1, '23:00'), (1, '00:00'), (1, '01:00'), (1, '02:00'), (1, '03:00'), (1, '04:00'), (1, '05:00'), (1, '06:00'), (1, '07:00'), (1, '08:00'), (1, '09:00'), (1, '10:00'), (1, '11:00'), (1, '12:00'), (1, '13:00'), (1, '14:00'), (1, '15:00'), (1, '16:00'), (1, '17:00'), (2, '18:00'), (1, '19:00'), (1, '20:00'), (1, '21:00'), (1, '22:00'), (1, '23:00')] \\
    
    Nearby roads: \\
    road 33889: speed\_limit 4, congestion\_level: [(1, '06:00'), ..., (2, '06:55')] \\
    ...\\
    
    Connectivity of roads: \\
    \texttt{[(23216, 27371, 500m), ..., (27371, 27370, 500m)]} \\
    
    \textbf{Question} \\[6pt]
    Based on the data, identify one of the typical time periods in the past 3 days when the traffic of the target road 27303 reaches its peak. \\
    A. 18:00-20:00 \\
    B. 00:00-02:00 \\
    C. 16:00-18:00 \\
    D. 06:00-08:00 \\
    
    \textbf{Note} \\[6pt]
    Please answer the question using the information given. Let’s work this out in a step-by-step way to be sure that we have the right answer. Then provide your final answer within the XML tag \texttt{<Answer>A/B/C/D</Answer>}. \\
    \bottomrule
  \end{tabularx}
  \vspace{-10pt}
  \label{tab:st_under_locex_q}
\end{table*}

\begin{figure}[t]
\centering
\includegraphics[width=\textwidth]{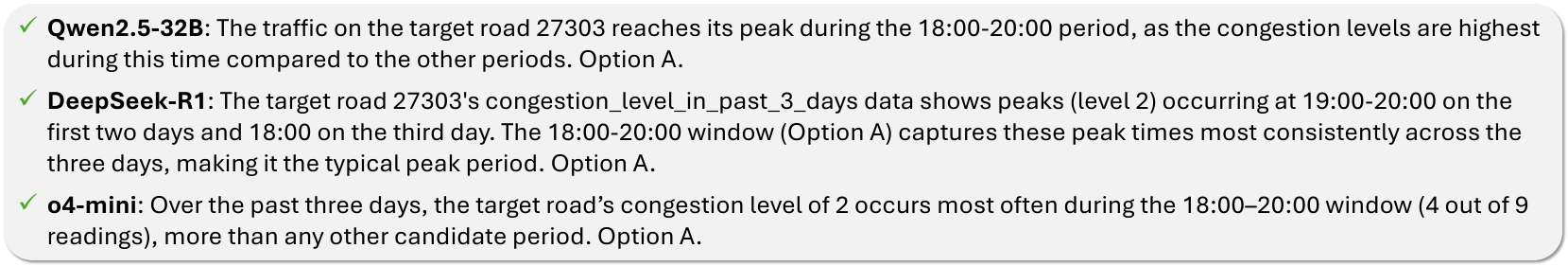}
\centering
\caption{The LLMs' reasoning on local extrema analysis.}
\label{fig:st_under_locex_a}
\vspace{-5pt}
\end{figure}

\begin{table*}[h]
\vspace{-5pt}
  \centering
  \caption{The spatiotemporal understanding QA for periodicity pattern analysis.}
  \vspace{-5pt}
  \begin{tabularx}{\textwidth}{X}
    \toprule
    \textbf{Task Description} \\[6pt]
    You are analyzing traffic congestion in the road network of Beijing, China. Given a target road and its nearby neighboring roads, along with time-stamped traffic conditions for each, perform a congestion analysis. \\
    \\
    \textbf{Data Schema} \\[6pt]
    - Target road: The road for which congestion patterns need to be analyzed. \\
    - Nearby roads: Roads within up to two hops from the target road. \\
    - speed\_limit: Speed limit on the road, with levels from 1 (lowest) to 4 (highest). \\
    - congestion\_level: A series of congestion levels (from 0 to 4, where 4 is the highest) recorded at specific times during the day. \\
    - congestion\_level\_in\_past\_3\_days: Historical time-series data of congestion levels over the past three days, recorded at specific time intervals. \\
    - Connectivity: A list of tuples in the format (road\_1, road\_2, distance), defining directed connections between roads with the specified distance in m. \\
    \\
    \textbf{Data} \\[6pt]
    Target road: \\
    road 21501: \\
    - speed\_limit: 4 \\
    - congestion\_level: [(2, '22:00'), (2, '22:05'), (2, '22:10'), (2, '22:15'), (1, '22:20'), (1, '22:25'), (1, '22:30'), (2, '22:35'), (2, '22:40'), (1, '22:45'), (1, '22:50'), (1, '22:55')] \\
    - congestion\_level\_in\_past\_3\_days: [(1, '00:00'), (1, '01:00'), (1, '02:00'), (1, '03:00'), (1, '04:00'), (1, '05:00'), (1, '06:00'), (1, '07:00'), (1, '08:00'), (1, '09:00'), (1, '10:00'), (1, '11:00'), (1, '12:00'), (1, '13:00'), (1, '14:00'), (1, '15:00'), (1, '16:00'), (2, '17:00'), (2, '18:00'), (2, '19:00'), (2, '20:00'), (2, '21:00'), (1, '22:00'), (1, '23:00'), (1, '00:00'), (1, '01:00'), (1, '02:00'), (1, '03:00'), (1, '04:00'), (1, '05:00'), (1, '06:00'), (2, '07:00'), (1, '08:00'), (1, '09:00'), (1, '10:00'), (1, '11:00'), (1, '12:00'), (1, '13:00'), (1, '14:00'), (1, '15:00'), (1, '16:00'), (2, '17:00'), (1, '18:00'), (2, '19:00'), (2, '20:00'), (2, '21:00'), (2, '22:00'), (2, '23:00'), (1, '00:00'), (1, '01:00'), (1, '02:00'), (1, '03:00'), (1, '04:00'), (1, '05:00'), (1, '06:00'), (2, '07:00'), (1, '08:00'), (1, '09:00'), (1, '10:00'), (1, '11:00'), (1, '12:00'), (1, '13:00'), (1, '14:00'), (1, '15:00'), (1, '16:00'), (2, '17:00'), (2, '18:00'), (2, '19:00'), (2, '20:00'), (2, '21:00'), (1, '22:00'), (1, '23:00')] \\[6pt]

    Nearby roads: \\
    ...

    Connectivity of roads: \\
    ...
    \\
    \textbf{Question} \\[6pt]
    Based on the data, identify the most possible periodicity of congestion changes on the target road 21501 over the past 3 days. \\
    A. 29 hours \\
    B. 17 hours \\
    C. 24 hours \\
    D. 32 hours \\

    \textbf{Note} \\[6pt]
    Please answer the question using the information given. Let’s work this out in a step-by-step way to be sure that we have the right answer. Then provide your final answer within the XML tag \texttt{<Answer>A/B/C/D</Answer>}. \\
    \bottomrule
  \end{tabularx}
  \vspace{-10pt}
  \label{tab:st_under_period_q}
\end{table*}

\begin{figure}[t]
\centering
\includegraphics[width=\textwidth]{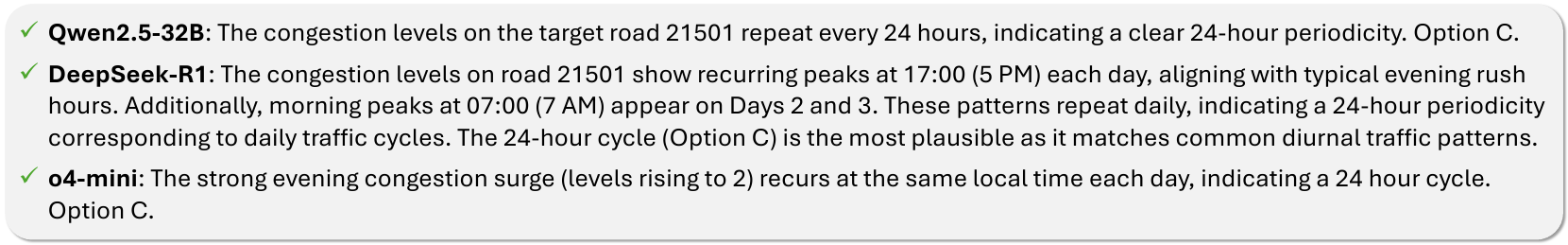}
\centering
\caption{The LLMs' reasoning on periodicity pattern analysis.}
\label{fig:st_under_period_a}
\vspace{-5pt}
\end{figure}

\begin{table*}[h]
\vspace{-5pt}
  \centering
  \caption{The forecasting QA.}
  \vspace{-5pt}
  \begin{tabularx}{\textwidth}{X}
    \toprule
    \textbf{Task Description} \\[6pt]
    You are analyzing traffic congestion in a road network of Beijing, China. Given a target road and its nearby neighboring roads, along with time-stamped traffic conditions for each, perform a congestion analysis. \\[6pt]

    \textbf{Data Schema} \\[6pt]
    - Target road: The road for which congestion patterns need to be analyzed. \\
    - Nearby roads: Roads within up to two hops from the target road. \\
    - speed\_limit: Speed limit on the road, with levels from 1 (lowest) to 4 (highest). \\
    - congestion\_level: A series of congestion levels (from 0 to 4, where 4 is the highest) recorded at specific times during the day. \\
    - congestion\_level\_in\_past\_3\_days: Historical time-series data of congestion levels over the past three days, recorded at specific time intervals. \\
    - Connectivity: A list of tuples in the format (road\_1, road\_2, distance), defining directed connections between roads with the specified distance in m. \\[6pt]

    \textbf{Data} \\[6pt]
    Target road: \\
    road 28340: \\
    - speed\_limit: 2 \\
    - congestion\_level: [(1, '19:00'), (1, '19:05'), (1, '19:10'), (1, '19:15'), (1, '19:20'), (1, '19:25'), (1, '19:30'), (1, '19:35'), (1, '19:40'), (1, '19:45'), (2, '19:50'), (2, '19:55')] \\
    - congestion\_level\_in\_past\_3\_days: [(1, '00:00'), (1, '01:00'), (1, '02:00'), (1, '03:00'), (1, '04:00'), (1, '05:00'), (1, '06:00'), (2, '07:00'), (3, '08:00'), (2, '09:00'), (2, '10:00'), (1, '11:00'), (1, '12:00'), (1, '13:00'), (2, '14:00'), (1, '15:00'), (1, '16:00'), (2, '17:00'), (2, '18:00'), (2, '19:00'), (1, '20:00'), (1, '21:00'), (1, '22:00'), (1, '23:00'), (1, '00:00'), (1, '01:00'), (1, '02:00'), (1, '03:00'), (1, '04:00'), (1, '05:00'), (1, '06:00'), (2, '07:00'), (3, '08:00'), (1, '09:00'), (1, '10:00'), (1, '11:00'), (1, '12:00'), (1, '13:00'), (2, '14:00'), (2, '15:00'), (1, '16:00'), (2, '17:00'), (2, '18:00'), (2, '19:00'), (1, '20:00'), (1, '21:00'), (1, '22:00'), (1, '23:00'), (1, '00:00'), (1, '01:00'), (1, '02:00'), (1, '03:00'), (1, '04:00'), (1, '05:00'), (1, '06:00'), (2, '07:00'), (2, '08:00'), (1, '09:00'), (1, '10:00'), (2, '11:00'), (1, '12:00'), (2, '13:00'), (2, '14:00'), (2, '15:00'), (2, '16:00'), (2, '17:00'), (1, '18:00'), (2, '19:00'), (1, '20:00'), (1, '21:00'), (1, '22:00'), (1, '23:00')] \\[6pt]

    Nearby roads: \\
    road 22668: \\
    - speed\_limit: 2 \\
    - congestion\_level: [(1, '19:00'), (1, '19:05'), (1, '19:10'), (1, '19:15'), (1, '19:20'), (1, '19:25'), (1, '19:30'), (1, '19:35'), (1, '19:40'), (1, '19:45'), (1, '19:50'), (1, '19:55')] \\

    road 27874: \\
    - speed\_limit: 2 \\
    - congestion\_level: [(1, '19:00'), (1, '19:05'), (1, '19:10'), (1, '19:15'), (1, '19:20'), (1, '19:25'), (1, '19:30'), (1, '19:35'), (1, '19:40'), (1, '19:45'), (1, '19:50'), (2, '19:55')] \\

    road 27875: \\
    - speed\_limit: 2 \\
    - congestion\_level: [(1, '19:00'), (1, '19:05'), (1, '19:10'), (1, '19:15'), (1, '19:20'), (1, '19:25'), (1, '19:30'), (1, '19:35'), (1, '19:40'), (1, '19:45'), (1, '19:50'), (2, '19:55')] \\[6pt]

    Connectivity of roads: \\
    \texttt{[(27874, 28340, 500m), (27875, 27874, 500m), (28340, 22668, 500m)]} \\

    \textbf{Question} \\[6pt]
    Based on the traffic conditions of the target road and its nearby roads, what is the most possible traffic congestion level of the target road 28340 at the next time step? \\
  \end{tabularx}
  \vspace{-10pt}
  \label{tab:forecast_q}
\end{table*}

\begin{table*}[h]
\vspace{-5pt}
  \centering
  \vspace{-5pt}
  \begin{tabularx}{\textwidth}{X}
    A. 1 \\
    B. 2 \\
    C. 3 \\
    D. 4 \\[6pt]

    \textbf{Note} \\[6pt]
    Let's solve this step by step. Finally, summarize your analysis, and provide your answer in JSON format, like: \\

    \texttt{\{} \\
    \quad\texttt{"summary": "YOUR\_SUMMARY",} \\
    \quad\texttt{"answer": "A/B/C/D"} \\
    \texttt{\}} \\
    \bottomrule
  \end{tabularx}
  \vspace{-10pt}
\end{table*}

\begin{figure}[t]
\centering
\includegraphics[width=\textwidth]{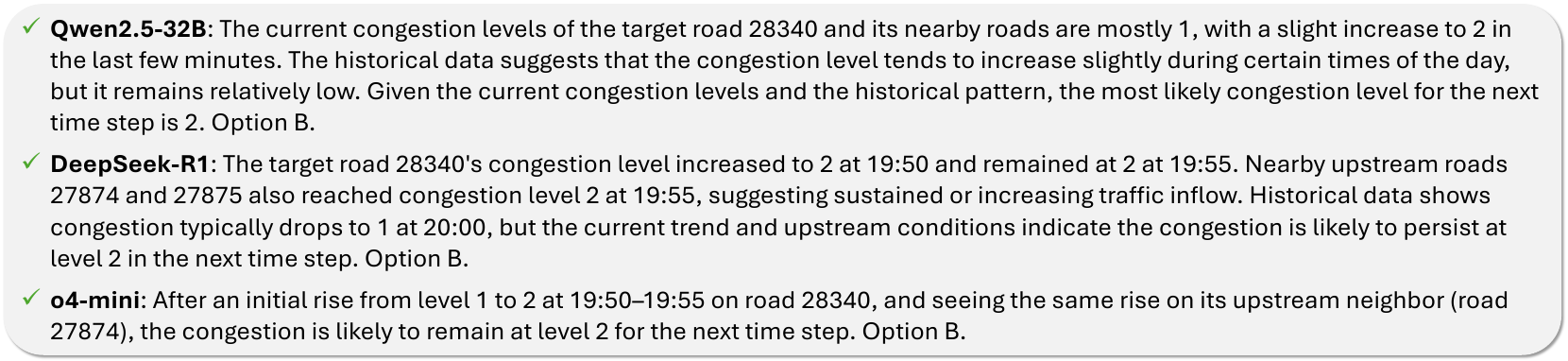}
\centering
\caption{The LLMs' reasoning on forecasting.}
\label{fig:st_under_forecast_a}
\vspace{-5pt}
\end{figure}

\begin{table*}[h]
\vspace{-5pt}
  \centering
  \caption{The reflection with environmental feedback QA.}
  \vspace{-5pt}
  \begin{tabularx}{\textwidth}{X}
    \toprule
    \textbf{Task Description} \\[6pt]
    Design a road network to connect unconnected regions to the existing connected regions in an urban area. Using the provided data, identify the roads that can be built to maximize connectivity. \\
    \\
    \textbf{Data Schema} \\[6pt]
    - connected\_regions: A list of regions that are already connected. \\
    - connectivity: A list of tuples in the format (node\_1, node\_2, distance), representing undirected roads between nodes with the specified distance (in km). \\
    - unconnected\_regions: A list of regions that are currently not connected to any other region. \\
    - region\_nodes: A list of boundary nodes defining the extent of each region. \\
    - available\_roads: A list of roads that can be built to improve connectivity. \\
    - road\_edge: A tuple in the format (node\_1, node\_2, distance), representing an undirected road that can be constructed between two nodes with the specified distance (in km). \\
    - new\_roads: A list of tuples in the format (node\_1, node\_2, distance), representing new roads that can be constructed after building the previous road.
    \\
    \textbf{Data} \\[6pt]
    connected\_regions: \\
    region 0: \\
    - region\_nodes: [3, 2, 1] \\
    region 26: \\
    - region\_nodes: [4, 2, 1, 14, 0, 13] \\
    ...
    connectivity: \\
    \texttt{[(node 1, node 3, 0.8km), (node 0, node 1, 0.17km), ..., (node 44, node 45, 0.36km)]} \\
    unconnected\_regions: \\
    region 4: \\
    - region\_nodes: [7, 9, 17, 5] \\
    region 15: \\
    - region\_nodes: [33, 26, 35, 31, 32] \\
    region 16: \\
    - region\_nodes: [17, 22, 5, 16, 21] \\
    ...\\
    available\_roads: \\
  \end{tabularx}
  \vspace{-10pt}
  \label{tab:plan_q}
\end{table*}

\begin{table*}[h]
\vspace{-5pt}
  \centering
  \vspace{-5pt}
  \begin{tabularx}{\textwidth}{X}
    road 7: \\
    - road\_edge: (node 3, node 5, 0.31km) \\
    - new\_roads: [(node 16, node 5, 0.31km)] \\
    road 20: \\
    - road\_edge: (node 17, node 9, 0.14km) \\
    - new\_roads: [(node 22, node 17, 0.17km)] \\
    road 30: \\
    - road\_edge: (node 15, node 19, 0.52km) \\
    - new\_roads: [(node 19, node 25, 0.24km)] \\
    ...\\
    \\
    \textbf{Question} \\[6pt]
    Based on the provided data, identify the most suitable road should be built next to connect the largest number of unconnected regions. \\
    A. road 7 \\
    B. road 72 \\
    C. road 84 \\
    D. road 81
    \\
    \textbf{Note} \\[6pt]
    Let's solve this step by step. Finally, summarize your analysis, and provide your answer in JSON format, like: \\
    \texttt{\{} \\
    \quad\texttt{"summary": "YOUR\_SUMMARY",} \\
    \quad\texttt{"answer": "A/B/C/D"} \\
    \texttt{\}} \\
    \bottomrule
  \end{tabularx}
  \vspace{-10pt}
\end{table*}

\begin{figure}[t]
\centering
\includegraphics[width=\textwidth]{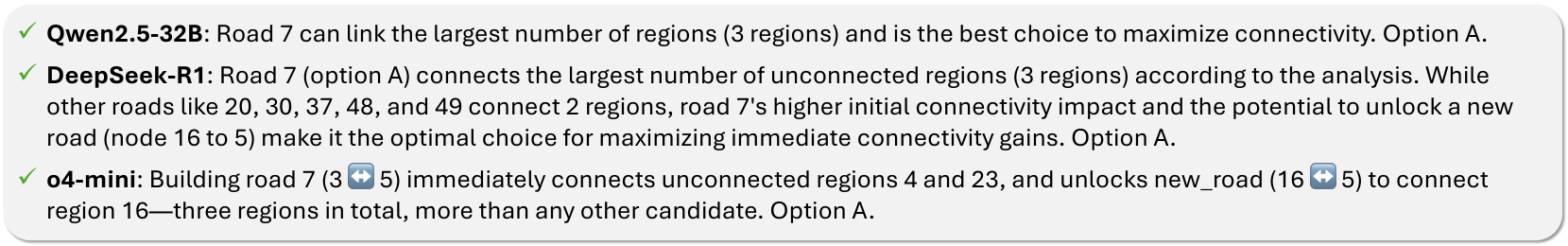}
\centering
\caption{The LLMs' reasoning on planning.}
\label{fig:st_under_plan_a}
\vspace{-5pt}
\end{figure}

\begin{table*}[h]
\vspace{-5pt}
  \centering
  \caption{The spatiotemporal understanding QA for periodicity pattern analysis.}
  \vspace{-5pt}
  \begin{tabularx}{\textwidth}{X}
    \toprule
    \textbf{Task Description} \\[6pt]
    We are tasked with determining the optimal location for a new electric vehicle (EV) charging station to maximize benefits by effectively meeting demand. The candidate locations, along with their expected charging station coverage, demand (number of vehicles), average distance, travel time, and waiting time to existing charging stations, are provided below. \\[12pt]

    \textbf{Data Schema} \\[6pt]
    - coverage: The number of locations that will be within the influence radius of the charging station after its construction. \\
    - demand: The number of vehicles requiring charging near the location. \\
    - distance: The average distance that vehicles near the location must travel to reach existing charging stations. \\
    - travel\_time: The average travel time for vehicles near the location to reach existing charging stations. \\
    - waiting\_time: The average waiting time at nearby existing charging stations. NaN indicates no station nearby and the highest waiting time. \\
    - charging\_time: The average charging time at nearby existing charging stations. NaN indicates no station nearby and the highest charging time. \\[12pt]

    \textbf{Data} \\[6pt]
    location 1: \\
    - coverage: 3.573 \\
    - demand: 22734.79 \\
    - distance: 284.90m \\
  \end{tabularx}
  \vspace{-10pt}
  \label{tab:ref_q}
\end{table*}

\begin{table*}[h]
\vspace{-5pt}
  \centering
  \vspace{-5pt}
  \begin{tabularx}{\textwidth}{X}
    - waiting\_time: 0.00h \\
    - charging\_time: 2.38h \\[6pt]
    location 2: \\
    ...\\
    location 3: \\
    ...\\
    location 4: \\
    ...\\

    \textbf{Analysis} \\[6pt]
    - Coverage ranking: \\
    location 1 (3.573) $>$ location 4 (3.572) $>$ location 3 (3.571) $>$ location 2 (3.571) \\[6pt]
    - Demand ranking: \\
    location 3 (121545.93) $>$ location 2 (81328.63) $>$ location 4 (27670.77) $>$ location 1 (22734.79) \\[6pt]
    - Distance ranking: \\
    location 1 (284.90m) $>$ location 4 (158.28m) $>$ location 3 (152.72m) $>$ location 2 (151.30m) \\[6pt]
    - Waiting time ranking: \\
    location 3 (0.04h) $>$ location 2 (0.04h) $>$ location 4 (0.01h) $>$ location 1 (0.00h) \\[6pt]
    - Charging time ranking: \\
    location 3 (21.38h) $>$ location 2 (17.03h) $>$ location 4 (8.51h) $>$ location 1 (2.38h) \\[12pt]

    \textbf{Question} \\[6pt]
    Evaluate each candidate location based on the provided data. Identify which location would deliver the most substantial benefits for an EV charging station, focusing on enhancing overall station coverage. \\
    A. location 1 \\
    B. location 2 \\
    C. location 3 \\
    D. location 4 \\[12pt]

    \textbf{Decision} \\[6pt]
    C \\[12pt]

    \textbf{Reason} \\[6pt]
    Location 3 offers the highest demand, which is critical for the success and utilization of the EV charging station, making it the most beneficial location despite location 1's higher coverage. \\[12pt]

    \textbf{Environment Feedback} \\[6pt]
    Station coverage increased by 181.05\% \\
    Charging time increased by 7.15\% \\
    Travel time increased by 59.22\% \\
    Waiting time increased by 83.68\% \\[12pt]
    \textbf{Note} \\[6pt]
    - Based on environment changes caused by the prediction, evaluate the correctness of the prediction and identify any weaknesses. \\
    - Verify if the Decision and its reasoning are accurate. \\
    - If the Decision is incorrect or inaccurate, propose a better alternative with a clear rationale. \\
    - Let's solve this step by step. Finally, summarize your analysis, and provide your answer in JSON format, like: \\[6pt]
    \texttt{\{} \\
    \quad \texttt{"summary": "YOUR\_SUMMARY",} \\
    \quad \texttt{"answer": "A/B/C/D"} \\
    \texttt{\}} \\
    \bottomrule
  \end{tabularx}
  \vspace{-10pt}
\end{table*}

\clearpage
\vspace*{-1cm} 
\begin{center}
\includegraphics[width=\textwidth]{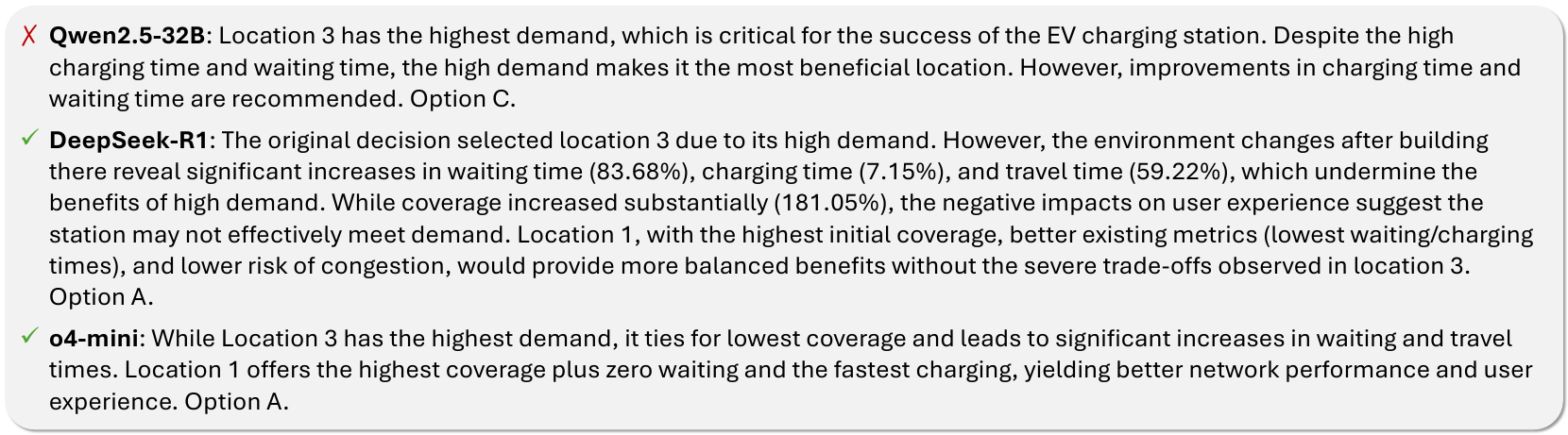}
\end{center}
\captionof{figure}{The LLMs' reasoning on reflection.}
\label{fig:st_under_ref_a}

\subsubsection{Repetition Issue}\label{subsubsec:agent_prompt}

\subsubsection{LLM Agent Prompt}\label{subsubsec:agent_prompt}

\begin{table*}[h]
\vspace{-5pt}
  \centering
  \caption{Task-solving prompt template.}
  \begin{tabularx}{\textwidth}{X}
    \toprule
    \textbf{Data Schema} \\[6pt]
    \texttt{<data\_schema>} \\[12pt]
    
    \textbf{Domain Knowledge} \\[6pt]
    \texttt{<domain\_knowledge>} \\[12pt]
    
    \textbf{Data} \\[6pt]
    \texttt{<data\_text>} \\[12pt]
    
    \textbf{Question} \\[6pt]
    Based on the provided task description, data schema, domain knowledge, provided data, and experience, you may first perform a spatiotemporal data analysis. Then, \texttt{<task\_target>}. \\[12pt]
    \\
    \textbf{Experience} \\[6pt]
    \texttt{<experience>} \\[12pt]
    
    \textbf{Note} \\[6pt]
    Let’s work this out in a step-by-step way to be sure that we have the right answer. Finally, provide your answer and summarize your analysis in JSON format, like: \\
    \texttt{```JSON} \\
    \texttt{\char123} \\
    \quad \texttt{"answer": "YOUR\_ANSWER",} \\
    \quad \texttt{"summary": "YOUR\_SUMMARY"} \\
    \texttt{\char125} \\
    \texttt{```} \\
    \bottomrule
  \end{tabularx}
  \vspace{-20pt}
  \label{tab:task_prompt}
\end{table*}

\begin{table*}[h]
  \centering
  \caption{Reflection prompt template.}
  \begin{tabularx}{\textwidth}{X}
    \toprule
    \textbf{Data Schema} \\[6pt]
    \texttt{<data\_schema>} \\[12pt]
    
    \textbf{Domain Knowledge} \\[6pt]
    \texttt{<domain\_knowledge>} \\[12pt]
    
    \textbf{Data} \\[6pt]
    \texttt{<data\_text>} \\[12pt]
    
    \textbf{Question} \\[6pt]
    Using the provided task description, data schema, domain knowledge, and data analysis, \texttt{<task\_target>}. \\[6pt]
    You need to: \\
    - Evaluate whether the previous \texttt{<decision\_or\_prediction>} was correct, based on the environmental feedback. \\
    - Identify any reasoning flaws or weaknesses in the original output. \\
    - If the \texttt{<decision\_or\_prediction>} was incorrect, propose a better alternative with clear justification. \\
    - Conclude with a high-level lesson learned. This summary must not include specific variable values or data IDs. \\[12pt]
    \\
    \textbf{Experience} \\[6pt]
    \texttt{<experience>} \\[12pt]
    
    \textbf{Previous Action} \\[6pt]
    \texttt{<decision\_or\_prediction>} \\[12pt]
    
    \textbf{Reason} \\[6pt]
    \texttt{<decision\_or\_prediction\_summary>} \\[12pt]
    
    \textbf{Environmental Feedback} \\[6pt]
    \texttt{<env\_feedback>} \\[12pt]
    
    \textbf{Note} \\[6pt]
    Let’s work this out in a step-by-step way to ensure we have the right answer. Finally, provide your reflection in JSON format as follows: \\
    \texttt{``` JSON} \\
    \texttt{\char123} \\
    \quad \texttt{"is\_correct": "YES/NO",} \\
    \quad \texttt{"answer": "YOUR\_ANSWER",} \\
    \quad \texttt{"experience": "YOUR\_SUMMARIZED\_EXPERIENCE"} \\
    \texttt{\char125} \\
    \texttt{```} \\
    \bottomrule
  \end{tabularx}
  \vspace{-20pt}
\end{table*}

\begin{table*}[h]
  \centering
  \caption{Summary prompt template.}
  \begin{tabularx}{\textwidth}{X}
    \toprule
    \textbf{Data Schema} \\[6pt]
    \texttt{<data\_schema>} \\[12pt]

    \textbf{Domain Knowledge} \\[6pt]
    \texttt{<domain\_knowledge>} \\[12pt]

    \textbf{Data} \\[6pt]
    \texttt{<data\_text>} \\[12pt]

    \textbf{Question} \\[6pt]
    Based on the task description, data schema, domain knowledge, and data sample, consolidate all experiences into \texttt{<memory\_num>} summarized experiences. \\[12pt]
    \\
    \textbf{Experience} \\[6pt]
    \texttt{<old\_experience>} \\
    \texttt{<new\_experience>} \\[12pt]

    \textbf{Note} \\[6pt]
    Let’s work this out step-by-step. Combine and restructure the information to create concise and meaningful summaries. Finally, provide your answer in JSON format as follows: \\
    
    \texttt{```JSON} \\
    \texttt{[} \\
    \quad \texttt{"YOUR\_EXPERIENCE\_1",} \\
    \quad \texttt{"YOUR\_EXPERIENCE\_2"} \\
    \texttt{]} \\
    \texttt{```} \\
    \bottomrule
  \end{tabularx}
  \vspace{-10pt}
  \label{tab:summary_prompt}
\end{table*}

\clearpage
\subsubsection{Spatiotemporal Understanding Post-Training Instruction Examples}

\begin{table*}[h]
  \centering
  \caption{The spatiotemporal understanding post-training instruction for distance analysis.}
  \vspace{-5pt}
  \begin{tabularx}{\textwidth}{X}
    \toprule
    \textbf{Task Description} \\[6pt]
    \texttt{A city is planning to increase green spaces and needs to consider distances from the new park to residential areas.} \\[12pt]
    \textbf{Data Schema} \\[6pt]
    \texttt{- distance: The distance to the park in km.} \\[12pt]
    \textbf{Data} \\[6pt]
    \texttt{neighborhood 0:} \\
    \texttt{- distance: 8.6km} \\
    \texttt{neighborhood 1:} \\
    \texttt{- distance: 8.89km} \\
    \texttt{neighborhood 2:} \\
    \texttt{- distance: 3.86km} \\
    \texttt{neighborhood 3:} \\
    \texttt{- distance: 16.19km} \\
    \texttt{neighborhood 4:} \\
    \texttt{- distance: 6.7km} \\
    \texttt{neighborhood 5:} \\
    \texttt{- distance: 7.96km} \\
    \texttt{neighborhood 6:} \\
    \texttt{- distance: 10.17km} \\
    \texttt{neighborhood 7:} \\
    \texttt{- distance: 13.97km} \\
    \texttt{neighborhood 8:} \\
    \texttt{- distance: 8.39km} \\
    \texttt{neighborhood 9:} \\
    \texttt{- distance: 7.08km} \\
    \texttt{neighborhood 10:} \\
    \texttt{- distance: 6.85km} \\[12pt]
    
    \textbf{Question} \\[6pt]
    How does neighborhood 9 compare to neighborhood 3? Which is closer to the park? \\[12pt]
    A. neighborhood 9 \\
    B. neighborhood 3 \\
    \\
    \textbf{Note} \\[6pt]
    Let’s work this out in a step-by-step way to be sure that we have the right answer. Finally, provide your answer and summarize your analysis in JSON format, like: \\
    \texttt{```JSON} \\
    \texttt{\char123} \\
    \quad \texttt{"answer": "A/B"} \\
    \texttt{\char125} \\
    \texttt{```} \\
    \\
    \textbf{Response} \\
    \texttt{<think>} \\
    Okay, let's see. The question is asking how neighborhood 9 compares to neighborhood 3 in terms of their distance to the new park, and which one is closer. The options are A for neighborhood 9 and B for neighborhood 3. \\
  \end{tabularx}
  \vspace{-60pt}
  \label{tab:distance_prompt}
\end{table*}

\begin{table*}[h]
  \centering
  \begin{tabularx}{\textwidth}{X}
    First, I need to look up the distances for both neighborhoods. From the data provided: \\
    - Neighborhood 3 has a distance of 16.19 km. \\
    - Neighborhood 9 has a distance of 7.08 km. \\
    
    So, comparing these two numbers, 7.08 km is definitely less than 16.19 km. That means neighborhood 9 is closer to the park than neighborhood 3. Therefore, the answer should be A. \\
    
    Wait, let me double-check to make sure I didn't mix up the numbers. Neighborhood 3 is over 16 km away, which is quite far, while neighborhood 9 is just over 7 km. Yep, that's correct. The smaller the distance number, the closer they are. So neighborhood 9 is closer. The answer is A. \\
    \texttt{</think>} \\[6pt]
    
    \texttt{\char123} \\
    \quad \texttt{"answer": "A"} \\
    \texttt{\char125} \\
    \bottomrule
  \end{tabularx}
  \vspace{-15pt}
\end{table*}

\begin{table*}[h]
\vspace{-10pt}
  \centering
  \caption{The spatiotemporal understanding post-training instruction for adjacency analysis.}
  \begin{tabularx}{\textwidth}{X}
    \toprule
    \textbf{Task Description} \\[6pt]
    \texttt{To enhance pedestrian safety, evaluate the number of crosswalks near schools and their relationship to traffic volume.} \\[12pt]
    \\
    \textbf{Data Schema} \\[6pt]
    \texttt{- connectivity: Each tuple (entity 1, entity 2, distance) defines a undirected edge from entity 1 to entity 2 with the specified distance in m.} \\[12pt]
    \\
    \textbf{Data} \\[6pt]
    \texttt{[(school 1, crosswalk 3, 1623m), (school 1, school 5, 98m), (school 1, school 4, 1428m), (crosswalk 2, school 4, 1995m)]} \\[12pt]
    \\
    \textbf{Question} \\[6pt]
    Among school 1 and crosswalk 3, which is further from school 5? \\
    \\
    \textbf{Note} \\[6pt]
    Let’s work this out in a step-by-step way to be sure that we have the right answer. Finally, summarize your analysis with JSON format, like: \\[6pt]
    \texttt{```JSON} \\
    \texttt{\char123} \\
    \quad \texttt{"summary": "YOUR\_SUMMARY"} \\
    \texttt{\char125} \\
    \texttt{```} \\
    \\
    \textbf{Response} \\
    \texttt{<think>} \\
    Okay, let's see. The task is to find which of school 1 or crosswalk 3 has a stronger spatial correlation with school 5. The data given is the connectivity between entities, which includes schools and crosswalks.
    
    First, the definition says that the connectivity is an undirected edge between two entities with a distance. So each tuple represents how those two entities are connected and the distance. I need to look at the data provided and see how school 5 is connected to these entities.
    
    Looking at the data entries:
    - (school 1, crosswalk 3, 1623m)
    - (school 1, school 5, 98m)
    - (school 1, school 4, 1428m)
    - (crosswalk 2, school 4, 1995m)
    
    School 5 is only mentioned once here, in the second tuple. So school 5 is directly connected to school 1 with a very short distance of 98 meters. That's really close. Now, crosswalk 3 is connected to school 1 but in the same tuple as school 1: (school 1, crosswalk 3, 1623m). So the distance here is 1623 meters, which is much farther.
    
    But how does crosswalk 3 relate to school 5? Since crosswalk 3 is connected to school 1, and school 1 is connected to school 5. So there's an indirect path from crosswalk 3 to school 5 through \\
  \end{tabularx}
  \vspace{-30pt}
  \label{tab:adjacency_prompt}
\end{table*}

\begin{table*}[h]
\vspace{-10pt}
  \centering
  \begin{tabularx}{\textwidth}{X}
    school 1. The total distance would be 1623m + 98m = 1721m. Or is it considered as the minimal path? Either way, compared to school 1's direct 98m to school 5, crosswalk 3 is much farther away. \\
    
    Spatial correlation here likely depends on proximity. The closer two entities are, the stronger their spatial correlation. So since school 1 is extremely close to school 5 (98m), while crosswalk 3 is over 1.6 km away from school 1 and hence over 1.7 km from school 5, school 1 would have a stronger spatial correlation with school 5.
    
    Crosswalk 3 is not directly connected to school 5 in the data. The only connection is through school 1. But even then, the total distance is way larger. The data doesn't have any other connections from crosswalk 3 to other entities except school 1. Therefore, school 1 is directly adjacent and very close, making it the stronger candidate.
    
    Wait, but the problem is about which (school 1 or crosswalk 3) is more correlated spatially with school 5. Since school 1 is directly connected with school 5 at 98m, that's a direct and short link. Crosswalk 3 is connected to school 1 (1623m away) which is connected to school 5 (98m). But crosswalk 3 itself isn't directly linked to school 5. So spatial correlation would depend on distance between entities. School 1 is right next to school 5, making their spatial correlation strong. Crosswalk 3 is much further away. So the answer should be school 1. \\[12pt]
    \texttt{</think>} \\
    \texttt{\char123} \\
    \quad \texttt{"answer": "school 1",} \\
    \quad \texttt{"summary": "School 1 has a stronger spatial correlation with school 5 because it is directly connected to school 5 via a very short edge (98 m). Crosswalk 3 is connected to school 1 but is much farther away (1623 m) and has no direct connection to school 5. Proximity strongly influences spatial correlation, and school 1's adjacency to school 5 outweighs crosswalk 3's indirect, distant relationship."} \\
    \texttt{\char125} \\
    \bottomrule
  \end{tabularx}
  \vspace{-10pt}
\end{table*}

\begin{table*}[h]
  \centering
  \caption{The spatiotemporal understanding post-training instruction for connectivity analysis.}
  \begin{tabularx}{\textwidth}{X}
    \toprule
    \textbf{Task Description} \\
    To assess the effectiveness of emergency response, evaluate the accessibility of fire stations to densely populated areas. \\
    \\
    \textbf{Data Schema} \\[6pt]
    \texttt{- connectivity: Each tuple (entity 1, entity 2, distance) defines an undirected edge from entity 1 to entity 2 with the specified distance in km.} \\[12pt]
    \\
    \textbf{Data} \\[6pt]
    \texttt{[(densely populated area 1, densely populated area 7, 5km), (densely populated area 1, fire station 3, 19km), (densely populated area 1, densely populated area 4, 6km), (densely populated area 1, densely populated area 6, 12km), (densely populated area 1, fire station 2, 4km), (fire station 2, densely populated area 5, 9km), (fire station 2, fire station 3, 1km), (fire station 2, densely populated area 6, 12km), (fire station 3, densely populated area 6, 10km), (fire station 3, densely populated area 4, 14km), (fire station 3, densely populated area 5, 8km), (fire station 3, densely populated area 7, 9km), (densely populated area 4, densely populated area 7, 15km), (densely populated area 5, densely populated area 7, 11km)]} \\[12pt]
    \\
    \textbf{Question} \\[6pt]
    Is fire station 2 reachable from densely populated area 5? \\[12pt]
    
    \textbf{Note} \\[6pt]
    Let’s work this out in a step-by-step way to be sure that we have the right answer. Finally, provide your answer and summarize your analysis in JSON format, like: \\
  \end{tabularx}
  \vspace{-10pt}
  \label{tab:connectivity_prompt}
\end{table*}

\begin{table*}[h]
  \centering
  \begin{tabularx}{\textwidth}{X}
  \texttt{```JSON} \\
    \texttt{\char123} \\
    \quad \texttt{"answer": "Yes/No",} \\
    \quad \texttt{"summary": "YOUR\_SUMMARY"} \\
    \texttt{\char125} \\
    \texttt{```} \\
    \texttt{<think>} \\
    Okay, let me try to figure out if fire station 2 is reachable from densely populated area 5. I need to see if there's a path connecting them through the given edges. Let's look at the data carefully. \\
    First, let's list out all the connections provided. Each tuple is an undirected edge, so movement can go both ways. The nodes include densely populated areas (like area1, area7, etc.) and fire stations (like fire station 2 and 3). \\
    Starting from densely populated area5. Let's see what connections it has. Looking through the list, I spot (fire station3, densely populated area5, 8km). Oh, right, area5 is connected to fire station3 with a distance. Additionally, there's (fire station2, densely populated area5, 9km). Wait, is that a direct connection between fire station2 and area5? Let me check again. Yes, the tuple (fire station2, densely populated area5, 9km) exists. So if area5 is directly connected to fire station2, then the answer should be yes. But wait, let me confirm if that connection is indeed there. \\
    Looking back at the data provided: \\
    The tuples are as follows: \\
    - (fire station2, densely populated area5, 9km) is one of them. So that means area5 is directly connected to fire station2 with 9km. Therefore, there's a direct path. No need to go through other nodes. Thus, area5 can reach fire station2 directly. So the answer should be Yes. \\
    \texttt{</think>} \\
    \texttt{\char123} \\
    \quad \texttt{"answer": "Yes",} \\
    \quad \texttt{"summary": "Densely populated area 5 is directly connected to fire station 2 via a 9km edge, as per the connectivity data provided."} \\
    \texttt{\char125} \\
    \bottomrule
  \end{tabularx}
  \vspace{-5pt}
\end{table*}

\begin{table*}[h]
  \centering
  \caption{The spatiotemporal understanding post-training instruction for duration analysis.}
  \begin{tabularx}{\textwidth}{X}
    \toprule
    \textbf{Task Description} \\[6pt]
    \texttt{Analyze the duration of power outages in industrial zones to develop strategies for improving grid reliability.} \\ [12pt]
    \\
    \textbf{Data Schema} \\ [6pt]
    \texttt{- duration: The duration of the power outage duration in h.} \\[12pt]
    \textbf{Data} \\[6pt]
    \texttt{power outage duration 0:} \\
    \texttt{- duration: 0.1h} \\
    \texttt{power outage duration 1:} \\
    \texttt{- duration: 0.53h} \\
    \texttt{power outage duration 2:} \\
    \texttt{- duration: 1.94h} \\
    \texttt{power outage duration 3:} \\
    \texttt{- duration: 0.82h} \\
    \texttt{power outage duration 4:} \\
    \texttt{- duration: 1.91h} \\
    \texttt{power outage duration 5:} \\
    \texttt{- duration: 1.18h} \\
    \texttt{power outage duration 6:} \\
    \texttt{- duration: 0.8h} \\[12pt]
    \textbf{Question} \\[6pt]
    Comparing power outage duration 5 and power outage duration 0, which is longer? \\[12pt]
  \end{tabularx}
  \vspace{-10pt}
  \label{tab:duration_prompt}
\end{table*}

\begin{table*}[h]
  \centering
  \vspace{-5pt}
  \begin{tabularx}{\textwidth}{X}
    \textbf{Note} \\[6pt]
    Let’s work this out in a step-by-step way to be sure that we have the right answer. Finally, provide your answer and summarize your analysis in JSON format, like: \\
    \texttt{```JSON} \\
    \texttt{\char123} \\
    \quad \texttt{"answer": "power outage duration 5/power outage duration 0"} \\
    \texttt{\char125} \\
    \texttt{```} \\
    \texttt{<think>} \\[6pt]
    Alright, let's tackle this problem step by step. I need to compare the duration of two specific power outages: number 5 and number 0. The data provided gives the durations in hours, so first I'll list those out. \\[6pt]
    Looking at the data schema, each entry is labeled as "power outage duration X" with a corresponding duration value. Specifically: \\[6pt]
    - Power outage duration 0 has a duration of 0.1 hours. 
    - Power outage duration 5 has a duration of 1.18 hours.
    The question asks which of these two is longer. So, I just need to compare the two numbers: 1.18h vs. 0.1h. \\[6pt]
    1.18 hours is clearly larger than 0.1 hours. Therefore, power outage duration 5 is longer than duration 0. I can confirm by converting to minutes if necessary. 0.1 hours is 6 minutes, while 1.18 hours is about 70.8 minutes. Yep, that's much longer. \\[6pt]
    No other data points affect this particular comparison since we are only looking at these two instances. So the answer is straightforward. \\[12pt]
    \texttt{</think>} \\[6pt]
    \texttt{\char123} \\
    \quad \texttt{"answer": "power outage duration 5"} \\
    \texttt{\char125} \\
    \bottomrule
  \end{tabularx}
  \vspace{-10pt}
\end{table*}

\begin{table*}[h]
\centering
\caption{The spatiotemporal understanding post-training instruction for chronology analysis.}
\vspace{-5pt}
\begin{tabularx}{\textwidth}{X}
\toprule
\textbf{Task Description} \\ [6pt]
You are tasked to analyze a user's historical POI stay records and visiting patterns. \\ [12pt]
\textbf{Data Schema} \\[6pt]
- event sequence: Each tuple (event, time, weekday) represents a event that happens at a specific time and day of the week. \\[12pt]
\\
\textbf{Data} \\[6pt]
The POI visiting sequence is below: \\
Day 1: \\
- trajectory: \texttt{[(Bike Shop 4355, Sunday, 06:00 PM), (Convenience Store 2809, Sunday, 07:14 PM)]} \\
- adjacency: N/A \\
Day 2: \\
- trajectory: \texttt{[(Post Office 37, Monday, 09:54 PM), (Bank 8352, Monday, 10:06 PM), (Food and Drink Shop 6831, Monday, 10:06 PM)]} \\ 
- adjacency: \texttt{[(Bank 8352, Post Office 37, 467.0m), (Bank 8352, Food and Drink Shop 6831, 117.0m)]} \\ 
Day 3:\\
- trajectory: \texttt{[(Convenience Store 2809, Tuesday, 08:15 PM), (Post Office 37, Tuesday, 08:15 PM), (Food and Drink Shop 1827, Tuesday, 08:15 PM)]} \\
- adjacency: \texttt{[(Post Office 37, Convenience Store 2809, 731.0m), (Post Office 37, Food and Drink Shop 1827, 471.0m)]} \\
Day 4:\\
- trajectory: \texttt{[(Convenience Store 2809, Thursday, 07:50 PM), (Post Office 37, Thursday, 07:58 PM), (Food and Drink Shop 1827, Thursday, 07:59 PM)]} \\
- adjacency: \texttt{[(Post Office 37, Convenience Store 2809, 731.0m), (Post Office 37, Food and Drink Shop 1827, 471.0m)]} \\
Day 5–30: ...\\
\end{tabularx}
\vspace{-10pt}
\label{tab:chronology_prompt}
\end{table*}

\begin{table*}[h]
\centering
\begin{tabularx}{\textwidth}{X}
\textbf{Question} \\[6pt]
Which event happened most recently? \\[12pt]

\textbf{Note} \\[6pt]
Let's work this out in a step-by-step way to be sure that we have the right answer. Finally, summarize your analysis with JSON format, like: \\
\texttt{\char123} \\
\texttt{```JSON} \\
\quad \texttt{"answer": "Train Station/Park/Convenience Store/Government Building/Other Great Outdoors/...",} \\
\quad \texttt{"summary": "YOUR\_SUMMARY"} \\
\texttt{\char125} \\
\texttt{```} \\
\\
\textbf{Response} \\
\texttt{<think>}\\
Okay, I need to figure out which event happens most recently based on the data provided. Let's look through all the days and their trajectory entries. The key here is to find the latest time on the latest date. \\[6pt]
First, I should check each day's trajectory to see the time and day of the week. All days are numbered from Day 1 to Day 30. Since each day is likely consecutive, Day 30 would be the last one in the sequence. Let me confirm with the data given.\\[6pt]
Looking at Day 30:\\
- The trajectory is [(Train Station 274, Sunday, 05:35 PM), (Miscellaneous Shop 13944, Sunday, 06:02 PM)]\\
So on Sunday at 06:02 PM, the user visited Miscellaneous Shop 13944. \\[6pt]
But wait, are there later days? No, the days go up to Day 30, so that's the latest day in the dataset. Now, looking at the entries on Day 30, the events are at 5:35 PM and 6:02 PM on Sunday.\\[6pt]
Are there any other entries in later days with a later time? Let me scan through the other days just to make sure. Let's check Day 29, 28, etc. to see if any times are later in the same day.\\[6pt]
Day 29 has a trajectory entry on Saturday at 04:58 AM. That's way earlier. Day 28: Friday, entries up to 07:03 PM. But Saturday would come before Sunday (the next day), so Day 30 is definitely after Day 28 and 29.\\[6pt]
Double-check the times on Day 30: the latest time there is 6:02 PM (18:02) on Sunday. Miscellaneous Shop 13944 is the last entry here.\\[6pt]
Are there any events on Day 30 that are later than 6:02 PM? The trajectory has two entries: 5:35 PM and 6:02 PM. The latest is 06:02 PM. That's the most recent time in the data.\\[6pt]
But wait, maybe another day with a more recent time? For example, maybe a day that's also Sunday but with a later time. Let me check all other Sundays.\\[6pt]
Looking through the data:\\[6pt]
Day 1 is Sunday (06:00 PM and 07:14 PM). Day 6 is Sunday (04:43 PM). Day 11: Sunday (07:20 PM). Day 16: Sunday (12:29 AM and 12:38 AM). Day 30: Sunday (05:35 PM, 06:02 PM). \\[6pt]
The latest Sunday in the dataset is Day 30, and the times there are 5:35 PM and 6:02 PM. But earlier, for example, Day 11 has an event at 7:20 PM. But that Day 11 is earlier (the 11th day), while Day 30 is the 30th day, so even though the time is earlier on a Sunday, the actual date (Day 30) is later. So the Day 30's 6:02 PM is more recent than any entries in previous Sundays, because Day 30 is a later day.\\[6pt]
Therefore, the most recent event is Miscellaneous Shop 13944 at 06:02 PM on Sunday. So the answer would be "Miscellaneous Shop". \\[6pt]
Wait, in the data entry for Day 30, each event in the trajectory is listed with time. Both entries are on Sunday. The Miscellaneous Shop is the later time. So yes, that's the most recent event in the entire dataset.\\
\texttt{</think>} \\
\texttt{\char123} \\
\quad \texttt{"answer": "Miscellaneous Shop"} \\
\texttt{\char125} \\
\bottomrule
\end{tabularx}
\vspace{-10pt}
\end{table*}

\begin{table*}[h]
  \centering
  \caption{The spatiotemporal understanding post-training instruction for trend analysis.}
  \begin{tabularx}{\textwidth}{X}
    \toprule
    \textbf{Task Description} \\[6pt]
    \texttt{You are tasked to analyze the time series data of solar energy generation power (in MW).} \\[12pt]
    \\
    \textbf{Data Schema} \\[6pt]
    \texttt{- time-series: Each tuple (value, time) represents a data reading in MW recorded at the specified time.} \\[12pt]
    \\
    \textbf{Data} \\[6pt]
    \texttt{[(30.5MW, 12:00:00), (30.25MW, 12:10:00), (29.8MW, 12:20:00), (29.7MW, 12:30:00), (29.45MW, 12:40:00), (28.2MW, 12:50:00), (27.05MW, 13:00:00), (26.65MW, 13:10:00), (26.4MW, 13:20:00), (25.9MW, 13:30:00), (25.35MW, 13:40:00), (24.6MW, 13:50:00)]} \\[12pt]
    \\
    \textbf{Question} \\[6pt]
    How is the trend of solar energy generation power (in MW)? First, examine the entire 12-step time series to determine the overall trend. If no clear trend emerges, divide the analysis into two segments of six steps each and assess trends separately for each half. \\[12pt]
    \\
    \textbf{Experience} \\[6pt]
    Let’s work this out in a step-by-step way to be sure that we have the right answer. \\[12pt]
    \textbf{Note} \\[6pt]
    Let’s work this out in a step-by-step way to be sure that we have the right answer. Finally, summarize your analysis with JSON format, like: \\
    \texttt{\char96\char96\char96 JSON} \\
    \texttt{\char123} \\
    \quad \texttt{"answer": "No trend/Increase/Decrease/No trend first, then increase/No trend first, then decrease/Increase first, then no trend/Decrease first, then no trend/Increase first, then decrease/Decrease first, then increase",} \\
    \quad \texttt{"summary": "YOUR\_SUMMARY"} \\
    \texttt{\char125} \\
    \texttt{\char96\char96\char96} \\
    \\
    \textbf{Response} \\
    \texttt{<think>} \\
    Okay, let's tackle this problem step by step. The task is to analyze the time series data of solar energy generation power in MW every 10 minutes from 12:00 to 13:50. The question is asking about the trend---whether it's increasing, decreasing, or if there's no clear trend. If the overall trend isn't clear, we need to split the data into two halves and analyze each separately. First, I'll look at the entire dataset. The timestamps start at 12:00 and go up to 13:50, making 12 data points, each 10 minutes apart. Let's list the values in order: 30.5, 30.25, 29.8, 29.7, 29.45, 28.2, 27.05, 26.65, 26.4, 25.9, 25.35, 24.6 MW. Visually examining these values, each subsequent reading is lower than the previous one. Let's check: From 30.5 to 30.25 (slight decrease), then 29.8, 29.7, 29.45... all the way down to 24.6. Each step shows a reduction. There's a consistent downward trend without any increases. So the overall trend might be a decrease. \\
    But wait, the instruction says to first check the entire 12-step series. Then, if no clear trend, split into two halves. Let me confirm each value again. Starting with 30.5 at 12:00, every next step is lower. For example: Step 1: 30.5 to 30.25 (down) Step 2: 30.5 \textrightarrow 30.25 \textrightarrow 29.8 \textrightarrow 29.7 \textrightarrow 29.45 \textrightarrow 28.2 (each subsequent is lower). Then from 28.2 to 27.05, down to 24.6. So, all 12 points are steadily decreasing. So the overall trend for the entire period is a decrease. \\
    Wait, no---I might need to check if the rate is consistent. For example, is the decrease steady across all data points? Let's list the values again: \\
  \end{tabularx}
  \vspace{-10pt}
  \label{tab:prompt}
\end{table*}

\begin{table*}[h]
  \centering
  \vspace{-5pt}
  \begin{tabularx}{\textwidth}{X}
    1. 30.5 2. 30.25 (-0.25) 3. 29.8 (-0.45) 4. 29.7 (-0.1) 5. 29.45 (-0.25) 6. 28.2 (-1.25) 7. 27.05 (-1.15) 8. 26.65 (-0.4) 9. 26.4 (-0.25) 10. 25.9 (-0.5) 11. 25.35 (-0.55) 12. 24.6 (-0.75) So the rate of decrease varies. For instance, steps 6 and 7 have larger drops (like -1.25 and -1.15). But even then, every step is lower than the preceding one. So regardless of how much it's decreasing, the trend is consistently downward. \\
    Wait, but the first half (first six steps) shows a downward trend, and the second half (last six steps) also shows a downward trend. So both halves individually have a decreasing trend. That means the entire time series has a consistent decreasing trend. Therefore, the overall answer should be a decrease. \\
    But wait---the problem says if the entire trend is unclear, split into two. But in this case, all points are decreasing. So the trend is clearly a decrease. Hence, the answer should be a single decreasing trend for the entire period. Therefore, the summary would state that the solar generation power shows a consistent decrease across all 12 steps with no periods of increase, leading to an overall downward trend. \\[12pt]
    \texttt{</think>} \\
    \texttt{\char123} \\
    \quad \texttt{"answer": "Decrease",} \\
    \quad \texttt{"summary": "The solar energy generation power shows a consistent decreasing trend across all 12 time steps. Starting at 30.5 MW at 12:00:00, the values decline steadily to 24.6 MW by 13:50:00, with each subsequent measurement being lower than the previous one. The rate of decrease varies slightly but remains unidirectional. Both the first and second halves of the data also exhibit decreasing trends, reinforcing the conclusion of an overall decline."} \\
    \texttt{\char125} \\
    \bottomrule
  \end{tabularx}
  \vspace{-5pt}
  \label{tab:prompt}
\end{table*}

\begin{table*}[h]
  \centering
  \caption{The spatiotemporal understanding post-training instruction for local extrema analysis.}
  \begin{tabularx}{\textwidth}{X}
    \toprule
    \textbf{Task Description} \\
    You are tasked to analyze the time series data of traffic flow. \\
    \\
    \textbf{Data Schema} \\[6pt]
    \texttt{%
    - time-series: Each tuple (value, time) represents a data reading recorded at the specified time.
    } \\[12pt]
    \\
    \textbf{Data} \\[6pt]
    \texttt{[(36.416666666666664, 2018-02-03 00:00:00), (19.833333333333332, 2018-02-03 01:00:00), (14.083333333333334, 2018-02-03 02:00:00), (19.333333333333332, 2018-02-03 03:00:00), (36.916666666666664, 2018-02-03 04:00:00), (67.08333333333333, 2018-02-03 05:00:00), (106.16666666666667, 2018-02-03 06:00:00), (161.75, 2018-02-03 07:00:00)...]} \\[12pt]
    \\
    \textbf{Question} \\[6pt]
    Identify one of the local maxima peak hours of traffic flow. \\[12pt]
    \textbf{Note} \\[6pt]
    Let’s work this out in a step-by-step way to be sure that we have the right answer. Finally, summarize your analysis with JSON format, like: \\
    \texttt{```JSON}
    \texttt{\char123} \\
    \quad \texttt{"summary": "YOUR\_SUMMARY"} \\
    \texttt{\char125} \\
    \texttt{```}
    \\
    \textbf{Response} \\
    \texttt{<think>} \\
    Okay, let's see. The task is to identify a local maxima peak hour in the traffic flow data provided. So first, I need to understand what a local maximum is in this context. A local maximum in a time series is a point where the value is higher than its immediate neighbors. That means, looking for \\
  \end{tabularx}
  \vspace{-10pt}
  \label{tab:prompt}
\end{table*}

\begin{table*}[h]
  \centering
  \begin{tabularx}{\textwidth}{X}
    times when the traffic flow increased to a peak and then started decreasing. Looking at the data provided, which spans from February 3rd to February 5th, each hour has a traffic value. So I need to go through each hour and check if the current hour's value is higher than the previous and next hour's values. Those would be the local maxima. \\
    Let me list out the values with their times. Starting with February 3rd: \\
    00:00: 36.416..., 01:00: 19.833... → decrease from previous., 02:00: 14.083..., 03:00:19.333..., 04:00:36.916..., 05:00:67.083..., 06:00:106.166..., 07:00:161.75..., 08:00:234.166..., 09:00:256.166..., 10:00:279.83..., 11:00:297.5..., 12:00:309.416..., 13:00:301.25 → This is lower than 12:00's 309.41, so 12:00 is a peak? Let's check. The 12:00 value (309.416) is higher than 11:00 (297.5) and then decreases at 13:00. Yes, so that's a local max. So 2018-02-03 12:00:00 could be a peak. \\
    Then 14:00:285.33, 15:00:276.16, 16:00:306.5 → comparing to 15:00 and 17:00. So 16:00's 306.5 vs previous 276.16 and next hour (17:00) 292.25. So 16:00 on the 3rd is higher than before and after, another local maximum. \\
    Then 17:00:292.25... then it goes down. Then 18th hour: decreases to 255.1, so the peak at 16:00. \\
    Looking at Feb 4th. Let me look through their hours. The data from Feb 4th starts at 00:00 again. Let's check peaks there. \\
    For example, at 08:00 Feb 4th: 200.58..., 09:00:239.66, 10:00:239.66, 11:00:272.58, 12:00:280.83, 13:00:283.91, 14:00:300.66. Wait, at 13:00, it's 283.91, then 14:00's 300.66. Is that higher than 15:00? Next hour (15:00 Feb4) is 256.25. So 14:00 is higher than both 13:00 and 15:00; so 14:00 on Feb4 is a local peak. \\
    Then at 19:00 Feb4, the value is 187.91, and 20:00 is 192.33 → not a peak. Let me check other times. \\
    On Feb 5th, the data shows: 05:00:288.41, 06:00:390.08..., 07:00:384.83, 08:00:366.16... So here, 06:00 is higher than 05:00 and 07:00. So that's a peak at 06:00 on Feb5th. Then later, at 17:00 Feb5: 392.16, which is higher than 16:00 (312.25) and 18:00 (305.0). So 17:00 on Feb5 is another local maximum. That's a notable peak. \\
    So possible candidates for local maxima are:\\
    - 2018-02-03 12:00:00 (309.41) \\
    - 2018-02-03 16:00:00 (306.5) \\
    - 2018-02-04 14:00:00 (300.66) \\
    - 2018-02-05 06:00:00 (390.08) \\
    - 2018-02-05 17:00:00 (392.16) \\
    Now, the question asks to identify one of these local maxima. The user just needs one, probably the highest or the most prominent. Let's check the values. The largest numbers I see are from February 5th. The 05:00 on Feb5 is 288, then 06:00 jumps to 390—almost 400, which is a significant spike. Then at 17:00 on Feb5, it goes up to 392.16. Comparing those two, 392.16 is higher than 390.08. Both are peaks. Let me confirm: \\
    05:00 Feb5:288.41 06:00:390.08 (peaks here, since the next hour at 07:00 is 384.83, which is lower) Then at 08:00:366.16 So at 06:00, higher than before and after — peak. \\
    Then 17:00 on Feb5 is 392.16, which follows 16:00 (312.25) and precedes 18:00 (305.0). So that's another peak. So 17:00 is higher than 06:00. Which one should I choose? \\
    The task is to find "one" of them. So selecting either would be correct. However, the highest local maximum would be 17:00 on Feb5 (392.16). Alternatively, the spike in 06:00 on Feb5 may be because it's part of the early morning rush, but the user just wants one example. Maybe the highest one. \\
    But let's verify if there are higher points anywhere else. \\
    Looking at Feb3: max is 309 at 12pm. Feb4's max at 14:00 is 300.6. So Feb5 peaks at 392 and 390. \\
    Yes. So I need to report one. For instance, 2018-02-05 17:00:00 with 392.16 as the local maximum, because that's higher than the surrounding hours. Alternatively, the 06:00 hour on Feb5 is also a local maximum. But since the task allows any one, it's sufficient to pick either. \\
    But the highest is 17:00 on the 5th. Alternatively, perhaps there is a typo, let me check the data again. \\
    Looking at February 5th's 05:00: 288.41. 06:00: 390.08. 07:00: 384.83. So 06:00 is higher than both, so that's a local maximum. Then again, at 17:00, 392.16 is higher than 16:00's 312 and 18:00's 305. So both those times are peaks. \\
    Of these, 392 at 17:00 is higher than the 390 at 6:00. So 17:00 is a higher peak. \\
  \end{tabularx}
  \vspace{-10pt}
  \label{tab:prompt}
\end{table*}

\begin{table*}[h]
  \centering
  \begin{tabularx}{\textwidth}{X}
    So the correct answer would likely be either of these. So to answer the user's question, I can choose the 17:00 on Feb5 as one example. \\[12pt]
    \texttt{</think>} \\
    \texttt{\char123} \\
    \quad \texttt{"answer": "2018-02-05 17:00:00 identified as a local maximum peak hour with traffic flow 392.16."}, \\
    \quad \texttt{"summary": "A significant local maximum peak occurs on 2018-02-05 at 17:00:00 with a traffic flow of 392.17. This hour is preceded by a rise from 312.25 (16:00) and followed by a drop to 305.0 (18:00), confirming it as a local peak within the time series."} \\
    \texttt{\char125} \\
    \bottomrule
  \end{tabularx}
  \vspace{-10pt}
  \label{tab:prompt}
\end{table*}

\begin{table*}[h]
  \centering
  \caption{The spatiotemporal understanding post-training instruction for periodicity analysis.}
  \vspace{-5pt}
  \begin{tabularx}{\textwidth}{X}
    \toprule
    \textbf{Task Description} \\
    You are tasked to analyze the time series data of electricity consumption (in Wh).\\[6pt]
    \\
    \textbf{Data Schema}\\
    - time-series: Each tuple (value, time) represents a data reading in Wh recorded at the specified time. \\[12pt]
    \\
    \textbf{Data} \\[6pt]
    The time-series of electricity consumption (in Wh) in every 1 hour is provided below:\\
    \texttt{[(112.0Wh, 2018-03-11 02:00:00), (114.0Wh, 2018-03-11 03:00:00), (116.0Wh, 2018-03-11 04:00:00), (114.0Wh, 2018-03-11 05:00:00), (115.0Wh, 2018-03-11 06:00:00), (122.0Wh, 2018-03-11 07:00:00), (136.0Wh, 2018-03-11 08:00:00)...]}
    \\[12pt]
    \\
    \textbf{Question} \\[6pt]
    How long is the period of electricity consumption (in Wh)? \\[12pt]
    
    \textbf{Note} \\[6pt]
    Let’s work this out in a step-by-step way to be sure that we have the right answer. Finally, summarize your analysis with JSON format, like: \\
    \texttt{```JSON}
    \texttt{\char123} \\
    \quad \texttt{"summary": "YOUR\_SUMMARY"} \\
    \texttt{\char125} \\
    \texttt{```} \\
    \texttt{<think>} \\
    Okay, so the task here is to determine the period of the electricity consumption based on the given time series data. Hmm, the term "period" in time series usually refers to the length of a repeating cycle or pattern. Let me think.

    Looking at the data, the timestamps start on 2018-03-11 at 02:00:00 and go through each hour until 2018-03-14 at 01:00:00. Let me first check the timestamps to see over how many days this data spans.
    
    Start date: March 11th at 2 AM. The data includes March 11, 12, 13, and part of the 14th (ending at 1 AM on the 14th). So that's roughly three days. Wait, from March 11th 02:00 to March 14th 01:00 is 24h * 3 days minus one hour. So about 71 hours? Let's count. From the first entry to the last entry: 11th 02:00 to 14th 01:00. The total number of data points is for each hour, and the entries seem consecutive with no gaps. Let me count the data points to check.
    
    Looking at the data provided, each entry increments by one hour. Let's see: first entry at 02:00, next at 03:00, up to the last at 01:00 on the 14th. The list starts with (11th 02:00) and ends with (14th 01:00). Since each entry is hourly, the total hours between first and last would be from 02:00 on 3/11 to 01:00 on 3/14. How many hours is that?
    
    From 3/11 02:00 to 3/14 02:00 would be exactly 3 days (72 hours). But since the last entry is 01:00 on 3/14, it's 72 -1 =71 hours. But each entry corresponds to an hour, so number of data points would be 71+1? Wait, no. Let me see. The first entry is at 02:00, next 03:00, etc. So from 02:00 to 24:00 on 3/11 is 22 entries (hours 2-23 inclusive, which is 22 hours?). Wait, no. Wait, \\
  \end{tabularx}
  \vspace{-10pt}
  \label{tab:periodicity_prompt}
\end{table*}

\clearpage
\vspace*{-1cm}
\noindent
\begin{minipage}{\textwidth}
  \centering
  \vspace{-5pt}
  \begin{tabularx}{\textwidth}{X}
    2 AM to 11 PM is 22 hours (since 24-2=22, but with each hour included. Wait, 02:00, 03:00,...23:00 for 3/11. Then next day starts at 00:00. Hmm, the data goes through 3/11, including the next day's entries. Let me recount the days covered. The data starts on 11th, March at 02:00 and ends on March 14th at 01:00. So 11th, 12th, 13th, and part of 14th.
    
    But the main point here isn't the span of the data but looking for the periodicity in the data. For electricity consumption, typically there's a daily pattern—higher during the day, lower at night. Let's check the values to see if they exhibit a daily cycle.
    
    Looking at the data: each day starts with lower values in the early morning, increasing during the day. For example, March 11th, around 10:00 AM (10:00:00), there's a spike to 543Wh, which seems like a huge jump compared to the previous hours. Then on March 12th, at 10:00, it's 526 Wh, also a spike. Same on March 13th: at 10:00, 552 Wh. So every day at around 10 AM, there's a big increase. After that, consumption remains high during the day and starts to drop in the evening. Then at 20:00 (8 PM) each day, it drops to around 359Wh (March 11), 361 (March 12), 357 (March 13), then drops further to around 160-168Wh in the late evening. Then this pattern repeats each day.
    
    So that's a clear daily period, where each day the consumption follows a similar pattern: low at night, peaks in the late morning/afternoon, then decreases again. So the period here is 24 hours—repeating every day.
    
    But wait the question asks "How long is the period of electricity consumption (in Wh)?" So it's looking for the duration of the cycle period, not the data's total duration but the cycle length. Given the daily repeats, the period is 24 hours.
    
    But need to confirm if the data indeed shows daily periodicity. Let me check across the days. Let's take the first three days.
    
    March 11th at 10:00: 543.0 Wh.
    
    March 12th 10:00:526.0.
    
    March13th 10:00:552.0.
    
    Then each day the peak hours of around 10:00-16:00 or so have high values, while nighttime is lower. Therefore, the repeating cycle is daily, which is 24 hours. Thus, the period is 24 hours. \\
    \texttt{</think>} \\
    \texttt{\{}\\
    \quad \texttt{"summary": "The electricity consumption exhibits a daily periodicity of 24 hours. Patterns repeat every day with lower consumption during nighttime and higher consumption during daytime, particularly spiking around late morning to afternoon hours. This daily cycle is consistent across all provided data points from March 11th to March 14th."} \\
    \texttt{\}} \\
    \bottomrule
  \end{tabularx}
  \vspace{-10pt}
  \label{tab:prompt}
\end{minipage}

\end{document}